%% file: tmi.tex
\documentclass[journal,twoside,web]{ieeecolor}
\usepackage{tmi}
\usepackage{cite}
\usepackage{amsmath,amssymb,amsfonts}
\usepackage{algorithmic}
\usepackage{graphicx}
\usepackage{textcomp}
\usepackage{hyperref}
\usepackage[capitalize]{cleveref}
\usepackage{mathtools}
\usepackage{svg}
\usepackage{tikz}
\usepackage{multirow}
\usepackage{array}
\usepackage{arydshln}
\usepackage{makecell}
\usepackage{xspace}
\usepackage{booktabs}
\usepackage[]{changes}

\definecolor{AddedColor}{RGB}{0,105,180}
\definecolor{DeletedColor}{gray}{0.5}
\definecolor{MarginColor}{rgb}{0.0,0.4,0.95}
\setaddedmarkup{%
  \textcolor{AddedColor}{#1}%
}
\setdeletedmarkup{%
  \textcolor{DeletedColor}{\sout{#1}}%
}
\setcommentmarkup{%
  \marginpar{\footnotesize\textcolor{MarginColor}{#1}}%
}
\newcommand{\rev}[3]{%
  \replaced[comment={#1}]{\mbox{}}{\mbox{}}%
  \replaced{#2}{#3}%
}

\renewcommand{\rev}[3]{#2}

\def\BibTeX{{\rm B\kern-.05em{\sc i\kern-.025em b}\kern-.08em
T\kern-.1667em\lower.7ex\hbox{E}\kern-.125emX}}
\begin{document}
\bstctlcite{IEEEexample:BSTcontrol}
\title{Unified and Semantically Grounded Domain Adaptation for Medical Image Segmentation}
\author{Xin Wang, Yin Guo, Jiamin Xia, Kaiyu Zhang, Niranjan Balu, Mahmud Mossa-Basha, {Linda Shapiro, \IEEEmembership{Fellow, IEEE}}, Chun Yuan
  \thanks{This work was supported by National Institute of Health (NIH)
  grants R01NS127317 and R01NS125635.}
  \thanks{Xin Wang is with the Department of Electrical and Computer Engineering, University of Washington, Seattle, WA 98195 USA (e-mail: xwang99@uw.edu). }
  \thanks{Yin Guo, Jiamin Xia and Kaiyu Zhang are with the Department of Bioengineering, University of Washington, Seattle, WA 98195 USA (e-mails: yinguo@uw.edu, jiaminx@uw.edu, kennyz@uw.edu). }
  \thanks{Niranjan Balu and Mahmud Mossa-Basha are with the Department of Radiology, University of Washington, Seattle, WA 98195 USA (e-mails: ninja@uw.edu, mmossab@uw.edu). }
  \thanks{Linda Shapiro is with the Department of Electrical and Computer Engineering and Paul G. Allen School of Computer Science and Engineering, University of Washington, Seattle, WA 98195 USA (e-mail: shapiro@uw.edu). }
  \thanks{Chun Yuan is with the Department of Radiology and Imaging Sciences, University of Utah, and the Department of Radiology, University of Washington, Seattle, WA 98195 USA (e-mail: cyuan@uw.edu). }
}

\maketitle

\begin{abstract}

  Most prior unsupervised domain adaptation approaches for medical image segmentation are narrowly tailored to either the source-accessible setting, where adaptation is guided by source-target alignment, or the source-free setting, which typically resorts to implicit adaptation mechanisms such as pseudo-labeling and network distillation. This substantial divergence in methodological designs between the two settings reveals an inherent flaw: the lack of an explicit, structured construction of anatomical knowledge that naturally generalizes across domains and settings.
  To bridge this longstanding divide, we introduce a unified, semantically grounded framework that supports both source-accessible and source-free adaptation.
  Fundamentally distinct from all prior works, our framework's adaptability emerges naturally as a direct consequence of the model architecture, without \rev{R3.2}{relying on explicit cross-domain alignment strategies}{the need for any handcrafted adaptation strategies}.
  Specifically, our model learns a domain-agnostic probabilistic manifold as a global space of anatomical regularities, mirroring how humans establish visual understanding. Thus, the structural content in each image can be interpreted as a canonical anatomy retrieved from the manifold and a spatial transformation capturing individual-specific geometry. This disentangled, interpretable formulation enables semantically meaningful prediction with intrinsic adaptability.
  Extensive experiments on challenging cardiac and abdominal datasets show that our framework achieves state-of-the-art results in both settings, with source-free performance closely approaching its source-accessible counterpart, a level of consistency rarely observed in prior works. Beyond quantitative improvement, we demonstrate strong interpretability of the proposed framework via manifold traversal for smooth shape manipulation. The results provide a principled foundation for anatomically informed, interpretable, and unified solutions for domain adaptation in medical imaging.
  \rev{R2.3}{The code is available at \href{https://github.com/wxdrizzle/remind}{https://github.com/wxdrizzle/remind}.}{}

  % We propose a unified and interpretable framework for unsupervised domain adaptation in medical image segmentation, which explicitly disentangles canonical anatomy and individual-specific geometry through a shared latent probabilistic manifold under the variational Bayesian framework. By leveraging a structured composition of learnable anatomical bases and a compact alignment mechanism over low-dimensional composition weights, our method enables efficient and explainable domain adaptation across both source-accessible and source-free settings, achieving state-of-the-art performance across multiple challenging benchmarks with a single model design. Moreover, we demonstrate that operating on semantically meaningful latent codes not only facilitates cross-domain alignment, but also yields anatomically plausible segmentations with improved interpretability, offering potential utility for downstream clinical analysis.

\end{abstract}


\input{body}

\bibliographystyle{IEEEtran}
\bibliography{_ref}
\end{document}

%% file: body.tex
\input{_symbols}

\section{Introduction}

 Learning-based methods have dominated medical image segmentation by enabling automatic and accurate delineation of anatomical structures, which facilitates a variety of clinical applications~\cite{ai_med_seg_survey}. This success, however, relies on large, well-annotated datasets that match the image characteristics of the intended domain. In practice, such data are often difficult to obtain due to variations in hardware, imaging protocols, patient populations, and disease manifestations. These domain shifts can severely degrade model performance, which has motivated unsupervised domain adaptation (UDA) to transfer knowledge from a labeled source domain to an unlabeled target domain with differing image characteristics, thereby alleviating the need for costly manual annotation of target data~\cite{DA_survey}.

Existing UDA methods primarily assume that source-domain data remain accessible during adaptation. In this \textit{source-accessible} setting, models can jointly utilize source and target data to learn domain-invariant representations. Numerous strategies have been explored, such as adversarial training, semi-supervised learning, and statistical alignment~\cite{ADVENT,MAPSeg,VarDA}. These methods often operate in high-dimensional feature spaces where alignment is computationally expensive and difficult to interpret. 
On the other hand, retaining source data during adaptation is not feasible in many real-world applications, due to privacy regulations, institutional policies, and data sharing restrictions. This gives rise to the \textit{source-free} setting, where only a pre-trained source model (but no source data) is available for target-domain adaptation, presenting greater challenges. Prior works for source-free adaptation frequently rely on complex self-training or entropy minimization techniques that are prone to instability, overfitting, and loss of anatomical fidelity~\cite{SFDA-DPL,tent,SFDA_media}. Crucially, existing source-accessible and source-free methods share common limitations: they lack explicit and explainable mechanisms to ensure that adapted features capture valid anatomical structures, which often results in implausible or fragmented segmentations.

This absence of explicit anatomical reasoning represents a fundamental inconsistency: surprisingly, despite \rev{R1.3}{the}{} trivial difference between the two settings, \ie, source-free simply requires completing source-domain learning before target-domain adaptation, existing literature has produced markedly different methodological designs. While source-accessible methods typically build upon domain alignment, source-free methods introduce entirely distinct pipelines based on self-supervision, pseudo-labeling, or distillation. This divergence in methodological design appears disproportionate to the underlying problem difference. 
% After all, both settings aim to transfer the segmentation knowledge learned from source samples to target-domain images. 
From a human perspective, adapting to a new domain, whether or not previous learning examples remain accessible, relies on the same conceptual understanding of anatomy~\cite{human_structure_mapping}. Therefore, we argue that the separation of current methods for the two settings reflects their inherent limitations, highlighting the need for a unified, interpretable framework that generalizes naturally across both scenarios.

In this work, we set out to close this gap, substantially extending our conference paper~\cite{remind}. Motivated by how humans adapt to unfamiliar imaging conditions, we propose a unified and semantically grounded Bayesian framework that applies seamlessly to both source-accessible and source-free adaptation.
Humans typically form a conceptual understanding of anatomy by memorizing typical shape patterns from labeled examples. When confronted with an unseen image, intuitively they 1) recall a representative shape and 2) deform it moderately to account for individual-specific structural variations~\cite{cognitive}. To emulate this process, we construct a low-dimensional, domain-agnostic latent probabilistic manifold, which encodes the full spectrum of representative structural patterns as weighted compositions of a few prototypical anatomical representations. Fine-grained geometric variations are then captured through additional spatial deformations.
This anatomy-aware manifold, shared across different domains and individual images, enables a clear disentanglement between domain-invariant canonical shape templates and individual-specific geometric details, which ensures structurally consistent and anatomically plausible predictions.
This design confers two key advantages. First, it enables adaptability to emerge naturally from the model architecture itself, without requiring \rev{R3.2}{explicit cross-domain alignment objectives}{any handcrafted adaptation objectives}. Second, the manifold serves as a compact memory that encapsulates anatomical priors, capable of adapting to target images whether or not source data are persistently available.

\begin{table}[t]
\centering
% \small
\setlength{\tabcolsep}{1.9pt}
\caption{\rev{}{Auditable comparison between the conference and journal variants under a unified evaluation pipeline. Values are mean $\pm$ std on the target-domain test set; brackets denote 95\% bootstrap confidence intervals.
For the conference version, the source-free (SF) setting is not applicable (N/A). $\Delta$ denotes the mean difference (Journal$-$Conference) under the source-accessible (SA) setting.}{}}
\label{tab:conf_vs_journal_auditable}
\begin{tabular}{l c c c c}
\toprule
{Dataset} & {Setting} & {Version} &
{DSC (\%) $\uparrow$} & {ASSD (mm) $\downarrow$} \\
\midrule

\multirow{4.5}{*}{MS-CMRSeg}
& \multirow{2}{*}{SA}
& Conf.  & 83.1$\pm$5.19 {\scriptsize [81.4, 84.7]} & 1.86$\pm$0.70 {\scriptsize [1.64, 2.08]} \\
&  & Jour. & 84.1$\pm$5.35 {\scriptsize [82.3, 85.7]} & 1.72$\pm$0.80 {\scriptsize [1.50, 2.00]} \\
\cmidrule(lr){2-5}
& \multirow{2}{*}{SF}
& Conf.  & N/A & N/A \\
&  & Jour. & 83.1$\pm$5.55 {\scriptsize [81.2, 84.7]} & 1.88$\pm$0.77 {\scriptsize [1.65, 2.14]} \\
% \specialrule{0.3pt}{2pt}{2pt}
\cmidrule(lr){1-5}
\multirow{4.5}{*}{AMOS22}
& \multirow{2}{*}{SA}
& Conf.  & 87.0$\pm$1.77 {\scriptsize [85.4, 88.6]} & 3.68$\pm$1.16 {\scriptsize [2.59, 4.63]} \\
&  & Jour. & 89.7$\pm$1.30 {\scriptsize [88.5, 90.8]} & 3.03$\pm$1.12 {\scriptsize [2.07, 3.97]} \\
\cmidrule(lr){2-5}
& \multirow{2}{*}{SF}
& Conf.  & N/A & N/A \\
&  & Jour. & 87.0$\pm$3.27 {\scriptsize [83.8, 89.3]} & 3.28$\pm$1.30 {\scriptsize [2.13, 4.44]} \\

\bottomrule
\end{tabular}

\vspace{0.35em}
\begin{tabular}{l c c}
\small
{SA $\Delta$ (Jour.$-$Conf.)} &
{$\Delta$DSC (\%)} & {$\Delta$ASSD (mm)} \\
\midrule
MS-CMRSeg & +1.04 & $-$0.13 \\
AMOS22    & +2.68 & $-$0.65 \\
\bottomrule
\end{tabular}
\end{table}

\rev{R3.1}{To make the empirical differences between this paper and the conference version explicit,
\cref{tab:conf_vs_journal_auditable} provides a side-by-side quantitative comparison under a unified
implementation and evaluation pipeline, covering both the source-accessible
and source-free settings.}{}
 
Our contributions can be summarized as follows:
\begin{enumerate}
    \item We propose a unified framework that seamlessly supports both source-accessible and source-free adaptation, achieving source-free performance closely matching its source-accessible counterpart.
    \item We introduce semantically grounded anatomical modeling, which emulates human visual understanding by explicitly disentangling canonical anatomy from individual geometry. This formulation leads to structurally consistent, robust, and interpretable predictions.
     \item Our method’s adaptability emerges naturally as an intrinsic property of the framework design. To the best of our knowledge, this is the first work to realize adaptation without \rev{R3.2}{explicit cross-domain alignment strategies}{any handcrafted strategies}.
\end{enumerate}

\section{Related Work}

\subsection{Unsupervised Domain Adaptation}

Previous source-accessible UDA works fundamentally relied on domain alignment through various strategies, such as adversarial networks, semi-supervised learning and statistical divergence. For example, ADVENT~\cite{ADVENT} and DARUNet~\cite{DARUNet} aligned the output or feature space by discriminators. MAPSeg~\cite{MAPSeg} leveraged masked auto-encoding to improve feature integrity. 
Generative approaches using variational inference, including VarDA~\cite{VarDA} and VAMCEI~\cite{VAMCEI}, derived statistical divergences between feature distributions. While these approaches captured domain-invariance, they often suffered from semantic ambiguity due to the absence of anatomical constraints, and required costly high-dimensional calculations.

Recent works have also investigated the more challenging source-free setting, relying on conventional strategies such as pseudo-labeling, entropy minimization, and distillation. For example, while Tent~\cite{tent} and AdaMI~\cite{SFDAIS} minimized prediction entropy, the latter also enforced class ratio priors. 
UPL-SFDA~\cite{upl-sfda} selected pseudo-labels via multiple prediction headers and refined results via dual-pass supervision. ProtoContra~\cite{ProtoContra} aligned target features to source model parameters and applied contrastive learning on unreliable samples. 
Despite their promising mitigation of prediction errors, existing source-free methods universally relied on refinement strategies that are vulnerable to noise in the outputs of source-trained models.

Few prior works have explored unsupervised adaptation from an anatomy-aware perspective. In contrast, our framework encapsulates structural priors from source data via a domain-agnostic manifold shared across all images. This formulation offers an interpretable means of preserving anatomical coherence, enabling adaptation to naturally emerge in both source-accessible and source-free settings.

\subsection{Variational Autoencoders in Medical Imaging}

Variational autoencoders (VAEs)~\cite{vae} provided a principled generative framework for effectively learning latent representations. 
For example, in image segmentation and synthesis, conditional VAEs were developed to disentangle anatomy from appearance, allowing representations to generalize across modalities~\cite{disentangled_vae}. Hierarchical VAEs were able to improve feature expressiveness through multiscale latent feature factorization in a theoretically-grounded probabilistic manner~\cite{nvae}.

In image registration, VAE-based models could effectively capture inter-modality anatomical consistency and learn diffeomorphic deformations. For example, BInGo explicitly disentangled anatomy and geometry for scalable and interpretable groupwise image registration~\cite{BInGo,bingo_journal} through unsupervised self-reconstruction. While powerful, learning domain-invariant structural features in this method relied on paired multi-modal images with a common anatomy, which is infeasible in domain adaptation since input images come from different subjects and spatial locations.

The proposed framework builds upon hierarchical and disentangled VAE formulations and introduces two key innovations: 1) a global latent space that \rev{R1.3}{is}{are} shared by all images and can generalize naturally across different domains and settings, 2) improving anatomical plausibility of segmentation predictions by encoding images as weighted combinations of anatomical representatives. This structured design greatly enhances interpretability and adaptability, while maintaining the flexibility and expressiveness of variational models.

\section{Methodology}

Let $(\mathcal{X},\mathcal{Y})$ denote the observation space, where $\mathcal{X} \subseteq \mathbb{R}^{D}$ is the image space, with $D$ the number of pixels, and $\mathcal{Y} \subseteq \{0, 1, \ldots, K\}^D$ is the segmentation label space, with $K$ the number of foreground classes.
The available data for learning include a labeled source dataset $\mathcal{D}^s = \{(\x_s^i, \y_s^i)\}_{i=1}^{N_s}$ sampled from a joint distribution $\mathbb{P}_s$ over $\mathcal{X} \times \mathcal{Y}$, and an unlabeled target dataset $\mathcal{D}^t = \{\x_t^i\}_{i=1}^{N_t}$ sampled from a marginal distribution $\mathbb{P}_t$ over $\mathcal{X}$.
% Considering the domain shift between $\mathbb{P}_s$ and $\mathbb{P}_t$, UDA aims to leverage both datasets to train a segmentation model that generalizes well to the target domain, without using target labels during training. 
We begin by introducing our framework from both theoretical and network implementation perspectives, and then describe how it is applied to both source-accessible and source-free settings.

\subsection{Disentangled Probabilistic Modeling}

\begin{figure}
  \centering
  \includegraphics[width=\linewidth]{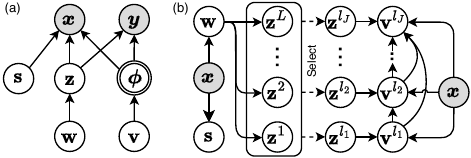}
  \caption{Graphical models of the proposed framework. (a) Generative model. (b) Inference model with hierarchical decomposition. Deterministic variables are in double circles, and observed variables are shaded. Dashed arrows denote selecting the subset $\{\z^{\lj}\}_{\lj\in\Lambda=\{l_1,\ldots,l_J\}}$.
   }
  \label{fig:graph}
\end{figure}

The generative structure of our model is presented in \cref{fig:graph}(a). Unlike prior methods that encoded all anatomical information into a single latent variable, we draw inspiration from how humans approach segmentation: intuitively, one first recalls a representative anatomical shape from prior knowledge, and then adapts it to the image through moderate spatial warping to account for individual-specific geometric details~\cite{cognitive}. Motivated by this perspective, we propose to explicitly disentangle the structural content of an image $\x$ into two distinct components: a canonical anatomical template $\z$ and a spatial deformation $\st$ that is parameterized by a stationary velocity field (SVF) $\v$ through $\st=\exp(\v)$~\cite{velocity}, such that $\x\circnew\st$ is spatially aligned to $\z$. 
This formulation leads to interpretable and geometry-aware representations of anatomical variations. In addition, we introduce a variable $\s$ to encode image style. 

While the template $\z$ encodes canonical anatomy, a single fixed $\z$ lacks expressiveness to account for topological diversity across images; on the other hand, extracting $\z$ from $\x$ without constraints may cause $\z$ to capture all anatomical variability and the deformation $\st$ to degenerate. To ensure both flexibility and disentanglement, we propose to condition $\z$ on a low-dimensional vector $\w\in\mathbb{R}^M$. This induces a structured latent manifold that supports interpretable and controllable shape retrieval, as detailed in \cref{sec:qzw,sec:interpret}. 

We leverage the variational Bayesian framework~\cite{vae} to effectively learn to infer the latent variables. Since $\x$ contains all information about the variables, we assume $\s$ is conditionally independent of all structure-related variables given $\x$. Thus, given a labeled source sample $(\x,\y)$, the joint and variational posterior distributions can respectively be written as 
\myeq{
&\pall\\
&=\pw \ps\pv\pz\px\py,\\
&\qall=\qs\qw\qz\qv.
}
Consequently, the evidence lower bound (ELBO) of the log-likelihood $\log p(\x,\y)$ is derived as
\myeq{
&\text{ELBO}\defas \E_{\qall}\left[\log \frac\pall \qall\right]\\
=&\E_{\qs\qw\qz\qv}\log\px &&\cdots\Lx\\
&+\E_{\qw\qz\qv}\log \py&&\cdots\Ly\\
&-\E_{\qw}\kl\qz\pz &&\cdots\Lz\\
&-\E_{\qw\qz}\kl\qv\pv &&\cdots\Lv\\
&-\kl\qs\ps-\kl\qw\pw,
}
where $D_\text{KL}$ is the Kullback-Leibler (KL) divergence. A similar decomposition applies to unlabeled target samples, with the omission of $\Ly$. The reconstruction probability $\px$ is modeled using pixel-wise Laplacian distributions, with their parameters predicted by a neural network. The segmentation term $\Ly$ is computed as the negative sum of the cross-entropy and Dice losses, following prior works~\cite{VarDA,VAMCEI}. Moreover, we assume deterministic posteriors for both $\w$ and $\s$, \ie, $\qw\defas\delta(\w-\wdeterm(\x))$, where $\delta$ is the Dirac delta and $\wdeterm$ is predicted from $\x$. As a result, inference yields $\w=\wdeterm$. The same formulation applies to $\s$. Therefore, the last two KL terms in the ELBO can be omitted during training.

We further factorize $\z,\v$ as $\z=(\z^l)_{l=1}^L$ and $\v=(\v^{l_j})_{\lj\in \Lambda}$~\cite{nvae,BInGo,bingo_journal,mri_cal_seg}, where $\Lambda=\{l_1,\ldots,l_J\}$ is a subsequence of $\{1,\ldots,L\}$, and a smaller $l$ indicates a coarser resolution. This hierarchical decomposition facilitates effective learning by capturing complex anatomical variability through progressively refined components. Thus, the spatial transformation is computed as $\st=\st^{l_1}\circnew\cdots\circnew\st^{l_J}$, with $\st^{l_{j}}$ parameterized by the velocity $\v^{l_j}$. We assume that different scales of $\z^l$ are independently conditioned on $\w$, and that $\v^\lj$ can be inferred given $\x$ and the template $\z^\lj$ at the same scale. In addition, we set $p(\v^\lj|\v^{<\lj})=p(\v^\lj)$ by design, where $<\lj$ denotes scales below $\lj$. Thus, $\Lz$ and $\Lv$ are simplified as
\myeq{
\label{eq:hierarchical_decomp}
\Lz=\E&_{\qw}\sum_{l=1}^L\kl \qzl \pzl,\\
\Lv=\E&_{\qw\qz}\sum_{\lj\in\Lambda}\E_{\qvlessl}\big(\\
&\kl\qvl\pvl\big).
}
where $q(\v^{<{l_1}}|\x,\z^{<{l_1}})\defas 1$. The inference structure of our model is thereby presented in \cref{fig:graph}(b). While each scale of $\z^l$ is computed independently, the inference of $\v^\lj$ proceeds in a coarse-to-fine manner: each scale depends on the template $\z^\lj$ and velocities $\v^{<\lj}$ inferred at coarser resolutions. This enables incrementally refining spatial transformations across scales. The KL terms in $\Lv$ are computed via the probabilistic SVF formulation~\cite{kl_v} to regularize $\st^{l_j}$ to be diffeomorphic. 

\subsection{Semantically Grounded Encoding with Shared Bases}
\label{sec:qzw}

Prior UDA works typically extract structural encodings directly from the input image. This conventional strategy often yields opaque and entangled representations, making interpretation and cross-domain adaptation difficult. In sharp contrast, we propose a structured extraction of the latent template $\z$, where its formation is not freely learned but explicitly modulated by the vector $\w$. To instill semantic organization and interpretability in the latent space, we introduce a small set of \textit{learnable} basis distributions $\{\qmzl\}_{m=1}^M$ for each scale $l$, \textit{shared across all images}, where $M$ is the length of $\w$. The posterior distribution of $\z^l$ is then modeled as a log-linear mixture~\cite{geo_mean} of the bases weighted by $\w$, \ie,
\myeq{
\qzl \propto \prod_{m=1}^M \left[\qmzl\right]^{w_m},
}
where we constrain $\w$ on the probability simplex $\Delta\defas\{\w\in\mathbb{R}^M|\w\succeq \mathbf{0},\mathbf{1}^\top\w=1\}$. This formulation restricts $q(\z^l|\w)$ to lie within the convex geometry spanned by the bases, introducing a strong inductive bias with several advantages:
\begin{enumerate}
    \item \textbf{Domain-agnostic regularization}: The domain-agnostic bases $\qmzl$ capture global anatomical regularities, enhancing robustness and generalizability.
    \item \textbf{Expressive shape composition}: The weight $\w$ blends prototypical structures encoded by bases, allowing the template $\z^l$ to express morphologically rich shapes via compositions of structural primitives.
    \item \textbf{Interpretable memory-like manifold}: The latent space is organized into a semantically structured manifold through the simplex-constrained $\w$, enabling interpretable traversal and feature extraction. This mechanism mimics how humans retrieve learned anatomical patterns from memory~\cite{human_vision}.
\end{enumerate}

Similar to the posterior, we define the prior as a uniform mixture over the bases, \ie,
$\pzl\defas p(\z^l)\propto\prod_{m=1}^M\left[\qmzl\right]^{1/M}$, without dependence on $\w$.
To allow for analytical tractability, we model each basis as a multivariate Gaussian, \ie, $\qmzl=\N {\mubold_m^l} {\Sigmabold_m^l}$, with diagonal covariance. Thus, we have $\qzl=\N{\mubold^l}{\Sigmabold^l}$, where 
\begin{equation}
  \label{eq:qz_calculation}
  \footnotesize{
\begin{alignedat}{2}
\Sigmabold^l=\left[\sum_{m=1}^Mw_m\left(\Sigmabold_m^l\right)^{-1}\right]^{-1},\
\mubold^l=\Sigmabold^l\sum_{m=1}^M\left[w_m\left(\Sigmabold_m^l\right)^{-1}\mubold_m^l\right].
  \end{alignedat}}
  \end{equation}
The same form applies to $\pzl$ with $\w=\frac1M\mathbf{1}$. 

Since $\Lz$ requires computing KL divergences for each image $\x$, we propose to replace it by the average of KLs over basis distributions, \ie,
\myeq{
\Lznew=\sum_{l=1}^L\frac{1}{M}\sum_{m=1}^M\kl{\qmzl}{p(\z^l)}.
}
This reduces the number of KL terms per batch from $LB$ to $LM$, where the batch size $B$ is typically much larger than $M$, substantially reducing computational cost. Despite the simplification, $\Lznew$ serves as an effective surrogate for $\Lz$: minimizing the former encourages each basis to stay close to the prior $\pzl$, which regularizes the mixture $\qzl$ to also remain close to the prior, thus effectively minimizing the original term $\Lz$.
The regularization $\Lznew$ can be interpreted as encouraging the bases to remain close to their average, and thus to each other. This constraint reduces the sensitivity of $\qzl$ to small perturbations in $\w$, facilitating more stable and reliable control of the latent template $\z^l$ via $\w$.

\begin{figure*}[t]
  \centering
  \includegraphics[width=\textwidth]{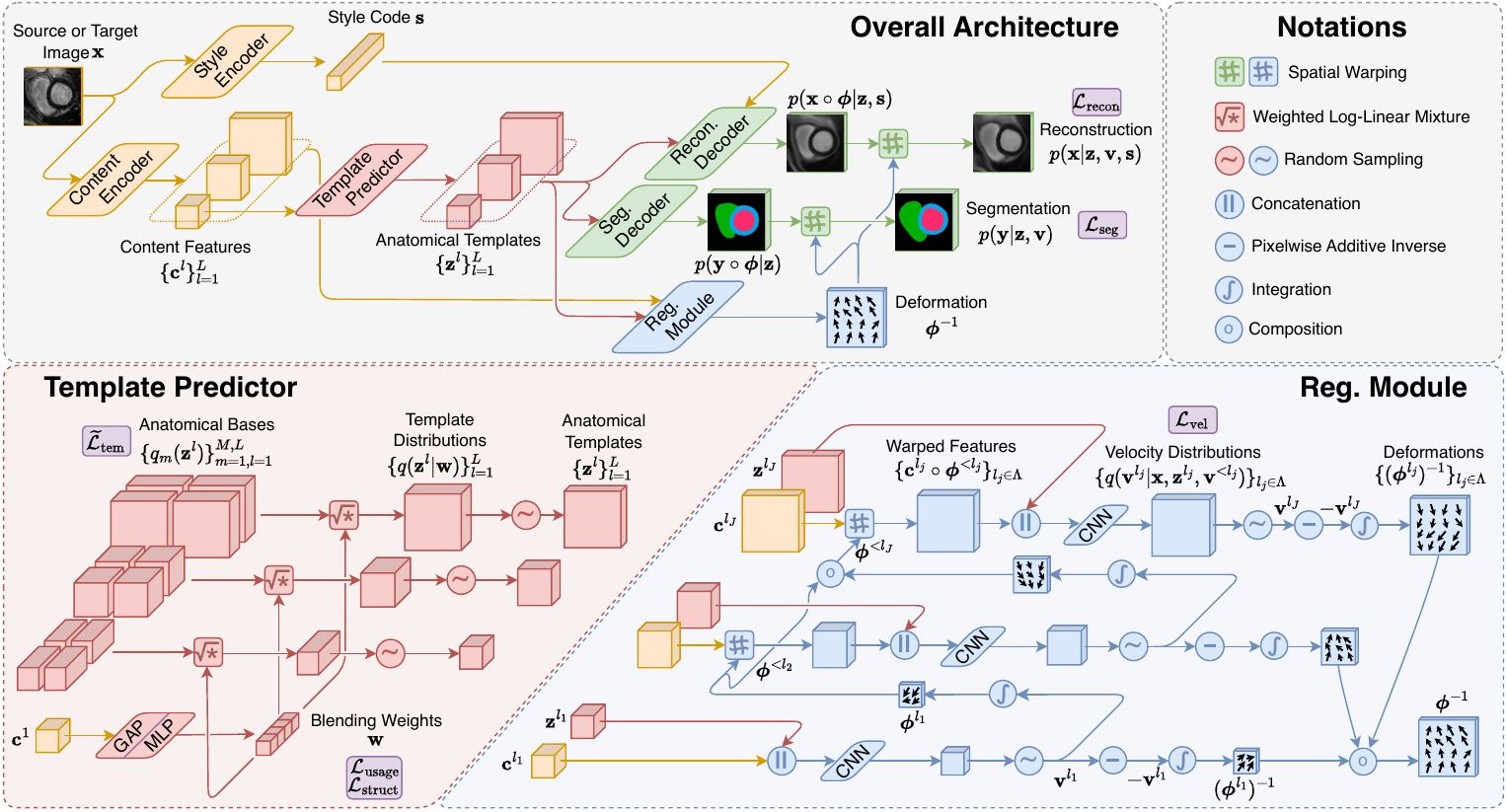}
  \caption{Network architecture for the proposed framework. Without loss of generality, the illustration utilizes $L=3$, $\Lambda=\{1,2,3\}$, and $M=4$. The Gaussian (\textit{resp.} Laplacian) distributions are represented by feature maps whose two halves of channels correspond to the mean and variance (\textit{resp.} scale), with the latter obtained via a Softplus function. Random samplings are performed during training, and replaced by taking the mathematical expectations during evaluation. The purple boxes correspond to the calculation of loss terms using related outputs. 
   }
  \label{fig:net_arch}
\end{figure*}

\subsection{Manifold Structuring for Emergent Adaptation}

A central premise of our framework is that canonical anatomical representations of all images are embedded into a shared, domain-agnostic manifold. In principle, this design allows domain adaptation to emerge naturally: the model only needs to interpret each image through this unified space, regardless of its domain origin. 
To fully realize this potential, we introduce two principled constraints that guide the internal organization of the manifold. These constraints ensure that the manifold is richly populated with structurally diverse and well-supervised representations, allowing target-domain inputs to be projected into regions with coherent anatomical semantics.

First, we promote balanced activation of anatomical bases within each source or target batch $\{\x_i\}_{i=1}^B$ through 
\myeq{
\Lbalance\defas \sum_{m=1}^M \max\left(0, \tau-\frac1{B}\sum_{i=1}^{B}w_m\left(\x_{i}\right)\right),
}
where $\tau=0.05$ defines a minimum usage threshold for each basis distribution. This encourages the model to utilize the full representational capacity of the manifold, which not only prevents basis underuse but also enhances the expressiveness and flexibility of the anatomical encoding.

Second, to ensure that the diversity in composition weights $\w$ reflect meaningful structural differences, we introduce a semantic dispersion constraint based on each labeled source batch $\{(\x_i^s,\y_i^s)\}_{i=1}^{B_s}$:
\myeq{
\Lwdice\defas \sum_{\substack{i,j=1\\i<j}}^{B_s}\left[\text{Sim}\left(\y_i^s\circnew\st_i^s,\y_j^s\circnew\st_j^s\right)-{C
\left(\w_i^s,\w_j^s
\right)}\right]^2
}
where $\text{Sim}$ denotes the Dice similarity between ground-truth segmentations warped by the inferred spatial deformations, and $C(\cdot,\cdot)$ is a similarity between their corresponding composition weights $\w$. To respect the non-Euclidean geometry of the probability simplex $\Delta$, we impose the Fisher-Rao metric $D_{\text{FR}}$ on the simplex to measure the distances among $\w$, \ie,
\myeq{
\fr\w{\w^\prime}\defas2\arccos\left(\sum_{m=1}^M \sqrt{w_mw_m^\prime}\right),\ \forall \w,\w^\prime\in\Delta,
}
which corresponds to the geodesic distance on a Riemannian statistical manifold~\cite{fisher-rao}. 
Therefore, $C(\cdot,\cdot)$ is calculated by transforming $D_{\text{FR}}$ into a bounded similarity score in [0,1], \ie, $C(\w,\w^\prime)\defas 1-{\fr\w{\w^\prime}}/\pi$.
The loss $\Lwdice$ explicitly associates latent diversity with anatomical diversity, thereby encouraging the manifold to be semantically organized and broadly covered by source-domain knowledge.

Together, these two constraints shape the manifold into an expressive, semantically coherent and fully utilized space, allowing adaptation to arise naturally through latent encoding, as a fundamental improvement over our conference paper~\cite{remind}  where adaptation relied on an explicit alignment loss. 
\rev{R4.7}{
Importantly, these regularizers do not introduce additional degrees of freedom.
Instead, they restrict the solution space of an otherwise underconstrained objective,
thereby reducing the risk of degenerate solutions rather than increasing overfitting.
For this reason, such regularization is particularly important in limited-data regimes,
where unsupervised adaptation is most prone to instability.}{}

\subsection{Network Architecture}

We design a specialized network architecture to support disentanglement of anatomy and geometry within the established probabilistic framework, as shown in \cref{fig:net_arch}.

\subsubsection{Inference of Anatomical Template}

A content encoder first extracts multiscale content features $\{\c^l\}_{l=1}^L$ from the input image. The coarsest feature map $\c^1$ is aggregated via global average pooling (GAP) and passed through a multilayer perceptron (MLP) followed by a Softmax to infer the composition weight $\w$. The anatomical template distributions $\qzl$ are then computed according to \cref{eq:qz_calculation}, leveraging a set of learnable feature maps that parameterize the basis distributions $\qmzl$. The templates $\z^l$ are then inferred based on $\qzl$.

\subsubsection{Inference of Velocity Fields}

Velocities and deformations are inferred via a dedicated registration module, conditioned on the content features $\c^\lj$ and anatomical templates $\z^\lj$. As indicated by the hierarchical factorization in \cref{eq:hierarchical_decomp}, the inference of each velocity $\v^\lj$ is conditioned on coarser-scale velocities $\v^{<\lj}$. Specifically, we first estimate the distribution $q(\v^{l_1}|\x,\z^{l_1})$ using a convolutional neural network (CNN) that takes $\c^{l_1}$ and $\z^{l_1}$ as input, yielding the initial velocity field $\v^{l_1}$ and corresponding deformation $\st^{l_1}$. For each scale $l_j > l_1$, we warp $\c^{l_j}$ by the composed deformation $\st^{<l_j} \defas \st^{l_{1}} \circnew \cdots \circnew \st^{l_{j-1}}$, resulting in a content representation partially aligned to the anatomical template. The warped feature $\c^{l_j}\circnew\st^{<{l_j}}$ and $\z^{l_j}$ are then fed into a CNN to estimate $\v^{l_j}$. After inferring all velocities $\{\v^\lj\}_{\lj\in\Lambda}$, the overall deformation $\st$ and its inverse $\st^{-1}$ are computed deterministically via integration~\cite{velocity} and composition.

\subsubsection{Segmentation and Reconstruction}

The anatomical template $\z=\{\z^l\}_{l=1}^L$ are fed into a segmentation decoder to generate a categorical probability map $p(\y\circnew\st|\z)$. This map is subsequently warped by the inverse transformation $\st^{-1}$ to yield the final segmentation prediction $\py$. For image reconstruction, a style encoder first extracts the style code $\s$ from the input image. Conditioned on both $z=\{\z^l\}_{l=1}^L$ and $\s$, a reconstruction decoder produces a Laplacian distribution $p(\x\circnew\st|\z,\s)$, which is then warped by $\st^{-1}$ to obtain the final reconstruction $\px$ of the input image.

\rev{R2.1}{We note that although the blending vector $\w$ is low-dimensional, it
parameterizes an anatomical manifold spanned by dense, pixel-wise
bases, from which full-resolution canonical segmentations are constructed.
The deformation $\st$ then specifies the pixel-to-pixel mapping from
canonical space to image space.
As a result, pixel-wise spatial detail is carried by dense bases
and deformation, rather than by $\w$ itself, and the proposed canonical-deformation
factorization does not constitute a spatial-bandwidth bottleneck for pixel-wise
segmentation prediction.
This representational design is further analyzed in the ablation study in
\cref{spatial_bottleneck_ablation}.}{}

\subsection{A Unified Paradigm for Source-Accessible and Source-Free Domain Adaptation}

We have established a theoretically and semantically grounded UDA framework through latent bases that encode and memorize global anatomical information. 
% The induced latent manifold serves as a geometry-aware scaffold that disentangles canonical anatomy, individual-specific geometry, and image style to facilitate interpretable, intrinsic alignment. 
We can now describe how our unified paradigm applies to the source-accessible and source-free settings, as a substantial extension of our conference paper~\cite{remind}. 

Formally, the ELBO has been derived as
\myeq{
\text{ELBO}^s(\x,\y)=&\lambda_1\Ly(\x,\y)+\lambda_2\Lx(\x)-\lambda_3\Lv(\x),\\
\text{ELBO}^t(\x)=&\lambda_2\Lx(\x)-\lambda_3\Lv(\x),
}
for a source or target observation, respectively, where $\lambda$'s are term weights, and $\Lznew$ has been separated out due to its independence from the data. We denote source and target batches as $\Bs\defas\{\x_i^s,\y_i^s\}_{i=1}^{B_s}$ and $\Bt\defas\{\x_i^t\}_{i=1}^{B_t}$, respectively, and the mean ELBO over a batch $\B$ as $\Lelbo(\B)$. In addition, we have introduced $\Lbalance,\Lwdice$ for manifold structuring. 

\subsubsection{Source-Accessible}

This conventional UDA setting involves a single-stage training using the source and target datasets $\mathcal{D}^s$ and $\mathcal{D}^t$ simultaneously. 
Thus, the overall loss is 
\myeq{
\mathcal{L}=&-\frac12\left[\Lelbo(\Bs)+\Lelbo(\Bt)\right]+\lambda_4\Lznew\\
&+\lambda_5\Lwdice(\Bs)+\frac12\left[\Lbalance(\Bs)+\Lbalance(\Bt)\right].
}

\subsubsection{Source-Free} In this setting, training is divided into two stages: the first has access only to the source data, while the second operates solely on the target domain. To this end, the loss for the first stage is
\myeq{
\mathcal{L}_1=-\Lelbo(\Bs)+\lambda_4\Lznew+\lambda_5\Lwdice(\Bs)+\Lbalance(\Bs).
}
After the first-stage training, the model has memorized structural patterns through the latent bases. Therefore in the second stage, we fix the bases $\qmzl$ and the segmentation decoder, and only optimize other parts of the network through
\myeq{
\mathcal{L}_2=-\Lelbo(\Bt)+\Lbalance(\Bt).
}

Note that $\mathcal{L}_1$ and $\mathcal{L}_2$ together constitute an exact decomposition of loss terms in $\mathcal{L}$. Thus, our formulation provides a unified and principled paradigm that seamlessly support\rev{R1.3}{s}{} both source-accessible and source-free settings.

\begin{table}[]
\caption{Batch sizes and loss weights for training our model.}
\label{tab:loss_weights}
\centering
\begin{tabular}{cccc}
\toprule
Dataset & Setting (Stage) & {Batch Size} & $\lambda_1$--$\lambda_5$ \\ \midrule
\multirow{3}{*}{MS-CMRSeg} & Source Access.     & 55   & 1,15,65,0.5,1      \\ 
& Source Free (1)      & 100        & 1,15,65,2,1          \\
&    Source Free (2)      & 70         & 0,15,65,0,0         \\ \midrule
\multirow{3}{*}{AMOS22}    & Source Access.     & 45         & 20,15,25,1e-4,10   \\ 
& Source Free (1)     & 80         & 20,15,25,1e-4,10      \\
&    Source Free (2)    & 60         & 0,15,25,0,0        \\ \bottomrule
\end{tabular}
\end{table}

\begin{table*}[t]
\caption{Comparison on the {\mscmr} dataset with state-of-the-art methods. \#Adapt denotes the number of adaptation strategies. In each setting, best results are marked in bold, and * indicates $p < 0.05$ (paired t-test) compared with Ours.}
\label{tab:result_sa_sf_mscmr}
\centering
\setlength{\tabcolsep}{4pt}
\begin{tabular}{clccccccccccccc}
\toprule
\multirow{2.5}{*}{Setting}&\multirow{2.5}{*}{Method}&\multirow{2.5}{*}{\#Adapt}&\multicolumn{4}{c}{DSC (\%) $\uparrow$} & & \multicolumn{4}{c}{ASSD (mm) $\downarrow$}\\
\cmidrule{4-7} \cmidrule{9-12}
 & && Average & Myo & LV & RV && Average & Myo & LV & RV  \\
\midrule
w/o Adapt. & Att-UNet~\cite{att_unet} & 0 & 54.2$\pm$10.2 & 48.3$\pm$10.3 & 70.7$\pm$12.6 & 43.5$\pm$13.2 & & 9.52$\pm$2.32 & 7.16$\pm$2.26 &8.07$\pm$3.43 &13.3$\pm$3.50 \\
\midrule
\multirow{5}{*}{\makecell{Source Access.}} & ADVENT~\cite{ADVENT} &3& 69.7$\pm$17.1$^*$ & 58.1$\pm$17.2 & 77.8$\pm$17.2 & 73.3$\pm$19.7 & & 3.75$\pm$3.38$^*$ & 4.05$\pm$7.12 &4.01$\pm$3.56 &3.19$\pm$2.06  \\
& VarDA~\cite{VarDA}  &1& 79.8$\pm$9.35$^*$ & 73.0$\pm$8.32 & 88.1$\pm$4.83 & 78.5$\pm$14.9 &  & 2.60$\pm$1.33$^*$ & 1.73$\pm$0.56 & 2.55$\pm$1.18 & 3.51$\pm$2.24\\
& DARUNet~\cite{DARUNet}  &7& 82.0$\pm$6.78$^*$ & 75.0$\pm$9.47 & 88.4$\pm$5.41 & 82.7$\pm$9.17 & & 2.26$\pm$1.05$^*$ & 1.64$\pm$0.71 &2.15$\pm$1.17 &2.99$\pm$1.97 \\
& MAPSeg~\cite{MAPSeg}  &3& 64.3$\pm$17.2$^*$ & 51.4$\pm$15.0 & 78.0$\pm$17.2 & 63.6$\pm$22.0 & & 5.17$\pm$4.76$^*$
 & 3.88$\pm$3.93 &5.56$\pm$6.28 &6.08$\pm$4.79 \\
& VAMCEI~\cite{VAMCEI}  &3& 82.5$\pm$5.39$^*$ & 75.8$\pm$6.64 & 88.2$\pm$5.27 & {83.6}$\pm${8.45} & &  1.93$\pm$0.89$^*$ & 1.54$\pm$0.70 &2.01$\pm$1.03 &\textbf{2.25}$\pm$\textbf{1.50}
\\
\midrule
\multirow{5}{*}{\makecell{Source Free}}& Tent~\cite{tent}         & 1&59.9$\pm$15.6$^*$ & 56.0$\pm$12.7 & 67.2$\pm$17.7 & 56.4$\pm$21.3 & & 9.58$\pm$3.20$^*$ & 6.62$\pm$2.48 &12.7$\pm$5.50 &9.43$\pm$3.86 \\ 
& FSM~\cite{FSM} & 5 & 67.8$\pm$15.0$^*$ & 61.0$\pm$13.8 & 77.4$\pm$14.1 & 65.2$\pm$22.5 & & 4.49$\pm$2.12$^*$ & 3.35$\pm$1.87 &4.54$\pm$2.75 &5.57$\pm$3.54\\

& AdaMI~\cite{SFDAIS}         &2& 71.6$\pm$8.06$^*$ & 68.7$\pm$9.90 & 84.9$\pm$6.38 & 61.0$\pm$11.3 & & 5.56$\pm$1.79$^*$ & 2.63$\pm$1.20 &4.71$\pm$2.82 &9.33$\pm$3.31  \\
& UPL~\cite{upl-sfda}  &2& 67.8$\pm$10.4$^*$ & 58.4$\pm$14.0 & 82.3$\pm$8.68 & 62.6$\pm$17.7 & & 6.13$\pm$2.68$^*$ & 4.39$\pm$2.02 &4.27$\pm$2.69 &9.75$\pm$6.58 \\
& ProtoContra~\cite{ProtoContra}  & 3&72.5$\pm$12.1$^*$ & 64.8$\pm$11.3 & 82.5$\pm$11.5 & 70.3$\pm$17.6 & & 4.07$\pm$1.99$^*$ & 2.69$\pm$1.38 &3.70$\pm$2.35 &5.81$\pm$3.76 \\
\midrule
Source Access. & \textbf{Ours} & 0&\textbf{84.1}$\pm$\textbf{5.35} & \textbf{78.5}$\pm$\textbf{5.16} & \textbf{90.0}$\pm$\textbf{4.28} & \textbf{83.9}$\pm$\textbf{8.64} & & \textbf{1.72}$\pm$\textbf{0.80} & \textbf{1.27}$\pm$\textbf{0.44} &\textbf{1.64}$\pm$\textbf{0.79} &2.26$\pm$1.53  
 \\
Source Free     &     \textbf{Ours}      & 0&\textbf{83.1}$\pm$\textbf{5.55} & \textbf{77.0}$\pm$\textbf{5.80} & \textbf{89.5}$\pm$\textbf{4.60} & \textbf{82.7}$\pm$\textbf{8.56} & & \textbf{1.88}$\pm$\textbf{0.77} & \textbf{1.38}$\pm$\textbf{0.48} &\textbf{1.78}$\pm$\textbf{0.87} &\textbf{2.49}$\pm$\textbf{1.42}
 \\
\bottomrule
\end{tabular}

\end{table*}

\begin{table*}[t]
\caption{Comparison on the {\amos} dataset with state-of-the-art methods. In each setting, best results are marked in bold, and * indicates $p < 0.05$ (paired t-test) compared with Ours.}
\label{tab:result_sa_sf_amos}
\centering
\setlength{\tabcolsep}{2.5pt}
\begin{tabular}{clcccccccccccc}
\toprule
\multirow{2.5}{*}{Setting}&\multirow{2.5}{*}{Method}&\multicolumn{5}{c}{DSC (\%) $\uparrow$} & & \multicolumn{5}{c}{ASSD (mm) $\downarrow$}\\
\cmidrule{3-7} \cmidrule{9-13}
  && Average & Liver & LK & RK & Spleen && Average & Liver & LK & RK & Spleen \\
\midrule
w/o Adapt. & Att-UNet & 9.38$\pm$8.67 & 29.3$\pm$25.3 & 0.17$\pm$0.35 & 4.85$\pm$6.06 & 3.17$\pm$5.91 & & 50.3$\pm$11.4 & 38.2$\pm$7.58 &49.1$\pm$10.5 &45.3$\pm$16.8 &68.7$\pm$19.7 \\
\midrule
\multirow{5}{*}{\makecell{Source Access.}} & ADVENT & 65.0$\pm$6.00$^*$ & 71.0$\pm$6.61 & 60.0$\pm$11.5 & 54.2$\pm$13.2 & 74.9$\pm$31.2 & & 6.70$\pm$1.78$^*$ & 6.11$\pm$3.11 &5.69$\pm$1.75 &9.01$\pm$2.18 &5.98$\pm$5.99 \\
& VarDA  & 81.4$\pm$4.73$^*$ & 85.2$\pm$6.70 & 79.8$\pm$8.86 & 78.4$\pm$6.76 & 82.0$\pm$7.47 & & 4.49$\pm$1.35$^*$ & 5.72$\pm$3.43 &5.08$\pm$1.19 &3.15$
\pm$0.65 &4.01$\pm$2.87\\
& DARUNet  & 85.8$\pm$5.20 & 87.1$\pm$5.56 & 82.1$\pm$9.46 & 82.4$\pm$12.8 & \textbf{91.6}$\pm$\textbf{3.85} & & 4.61$\pm$2.28 & 5.24$\pm$4.29 &5.89$\pm$3.15 &3.61$\pm$2.16 &3.70$\pm$2.50\\
& MAPSeg  & 88.5$\pm$3.02 & \textbf{92.2}$\pm$\textbf{2.45} & 84.0$\pm$5.62 & 87.9$\pm$4.38 & 90.0$\pm$8.10 & & 4.21$\pm$2.34 & 4.95$\pm$6.24 &5.15$\pm$2.52 &3.46$\pm$2.61 &3.26$\pm$3.24 \\
& VAMCEI  & 87.1$\pm$3.15 & 88.6$\pm$5.01 & 85.7$\pm$6.72 & 84.5$\pm$3.72 & 89.8$\pm$3.09 & & 3.53$\pm$1.92 & \textbf{3.64}$\pm$\textbf{1.90} &2.33$\pm$0.88 &5.07$\pm$3.72 &\textbf{3.10}$\pm$\textbf{2.34}\\
\midrule
\multirow{5}{*}{\makecell{Source Free}}& Tent         & 25.7$\pm$25.0$^*$ & 17.2$\pm$26.8 & 12.2$\pm$24.3 & 18.3$\pm$35.4 & 54.9$\pm$28.9 & & 35.3$\pm$12.6$^*$ & 33.0$\pm$16.7 &49.2$\pm$12.7 &44.6$\pm$20.5 &14.4$\pm$5.53\\ 
& FSM & 2.57$\pm$5.09$^*$ & 0.00$\pm$0.00 & 0.30$\pm$0.59 & 0.71$\pm$1.42 & 9.28$\pm$18.4 & & 70.1$\pm$21.4$^*$ & 88.4$\pm$23.8 &42.7$\pm$13.4 &78.4$\pm$32.3 
&71.2$\pm$29.1
\\
& AdaMI         & 70.6$\pm$2.92$^*$ & 85.6$\pm$4.35 & 71.5$\pm$7.56 & 37.2$\pm$4.37 & 88.1$\pm$6.70 & & 14.4$\pm$3.97$^*$ & 5.42$\pm$2.13 &18.2$\pm$8.69 &29.7$\pm$4.49 &4.11$\pm$2.75\\
& UPL  & 44.4$\pm$18.6$^*$ & 75.9$\pm$7.51 & 17.5$\pm$35.0 & 12.2$\pm$22.1 & 71.9$\pm$36.0 & & 31.2$\pm$7.71$^*$ & 17.4$\pm$9.01 &40.6$\pm$11.5 &50.3$\pm$15.7 &16.4$\pm$23.1\\
& ProtoContra  & 76.4$\pm$8.26$^*$ & 81.9$\pm$21.9 & 57.1$\pm$20.0 & 78.3$\pm$5.37 & 88.2$\pm$4.30 & & 11.0$\pm$2.92$^*$ & 5.20$\pm$5.76 &18.1$\pm$8.26 &16.0$\pm$5.16 &4.64$\pm$4.04\\
\midrule
Source Access. & {\textbf{Ours}} & \textbf{89.7}$\pm$\textbf{1.30} & 89.4$\pm$5.22 & \textbf{90.5}$\pm$\textbf{1.85} & \textbf{89.5}$\pm$\textbf{1.55} & 89.3$\pm$3.37 & & \textbf{3.03}$\pm$\textbf{1.12} & 4.33$\pm$2.10 &\textbf{1.33}$\pm$\textbf{0.17} &\textbf{1.36}$\pm$\textbf{0.27} &
5.11$\pm$2.74
 \\
Source Free       &     \textbf{Ours}     &\textbf{87.0}$\pm$\textbf{3.27} & \textbf{87.7}$\pm$\textbf{6.39} & \textbf{85.9}$\pm$\textbf{4.98} & \textbf{86.1}$\pm$\textbf{5.50} & \textbf{88.2}$\pm$\textbf{3.96} & & \textbf{3.28}$\pm$\textbf{1.30} & \textbf{4.36}$\pm$\textbf{2.04} &\textbf{2.64}$\pm$\textbf{1.06} &\textbf{2.64}$\pm$\textbf{1.43} &\textbf{3.50}$\pm$\textbf{2.43} 
 \\
\bottomrule
\end{tabular}
\end{table*}

\section{Experiments and Results}
\subsection{Datasets}
We conducted comprehensive experiments on two public datasets across the source-accessible and source-free settings to evaluate the accuracy, robustness, and interpretability of the proposed framework. The datasets encompass a wide spectrum of imaging characteristics, including various organs, modalities, protocols, pathologies, and populations. 
\subsubsection{{\mscmr}}
The MS-CMRSeg 2019 challenge~\cite{mscmr} provides cardiac MRI scans in three sequences, bSSFP, LGE and T2, acquired from 45 patients. Ground-truth segmentations are available for the left ventricle (LV), right ventricle (RV), and myocardium (Myo). 
\rev{R2.2}{We}{Following prior studies [5],[13], we} designated 35 bSSFP subjects as the source domain and 45 LGE subjects as the target domain, \rev{}{where}{allocating} 5 LGE subjects \rev{}{are used}{} for validation \rev{}{and model selection}{}, and the remaining 40 \rev{}{LGE}{} subjects \rev{}{are used}{} for \rev{}{final}{} testing. \rev{}{During training, images from all 45 LGE subjects are used without accessing any target-domain labels.}{} 
To simulate an unpaired scenario, the 2D slices were randomly shuffled \rev{}{after the subject-wise split has been fixed. The protocol above is identical to that used in prior studies~\cite{VarDA,VAMCEI}}{}.
All images were resampled to an in-plane resolution of 0.76 mm, cropped to $192\times192$ pixels to standardize the field of view, and min-max normalized.
\textit{The major challenges of this dataset include 1) limited number of training slices, and 2) considerable domain shifts due to complex intensity patterns, imaging noise, artifacts, weak contrasts across substructures, and shape deformation induced by pathology (\eg, scarring).}

\subsubsection{{\amos}}
The AMOS 2022 challenge~\cite{amos22} provides a multi-center collection of unpaired abdominal CT and MRI scans, encompassing multiple diseases and imaging protocols. In this study, we focused on segmenting four key organs: liver, spleen, left kidney (LK), and right kidney (RK). Following previous works~\cite{SIFA}, we selected 25 MRI scans as the source domain and 35 CT scans as the target domain. The CT data were further split into 25 scans for training, 5 for validation, and 5 for testing\rev{R2.2}{, with all splits being subject-wise disjoint}{}. To ensure consistency across samples, axial slices were extracted from the 3D volumes, resampled to a uniform spacing of 1.5 mm, and cropped to a consistent field of view centered on the organs of interest. Pixel values were clipped to $[0,250]$ for CT and the 0--99.5 percentile range for MRI, and then min-max normalized.
\textit{The major challenges of this dataset include large structural differences across images and domain shifts due to multi-modality, noise and artifacts.}

\subsection{Experimental Setups}
\subsubsection{Implementation Details}
We set the number of hierarchical levels to $L=5$ with $\Lambda=\{1,3,5\}$, and the number of basis distributions $M$ to be 6 and 10 for the {\mscmr} and {\amos} datasets, respectively. The content encoder used an attention U-Net~\cite{att_unet}, with the outputs from attention layers serving as multi-scale content features $\c^l$ of the input image. The style encoder is a Conv-LeakyReLU-AvgPool-Linear sequence, producing a 128-dimensional style code $\s$. While the reconstruction and segmentation decoders do not share parameters, they both adopted the U-Net decoding structure that takes multi-scale feature maps as input. The reconstruction decoder also incorporated adaptive instance normalization~\cite{AdaIN} to modulate the output using the style code. Each CNN in the registration module comprised four Conv-Norm-LeakyReLU sequences followed by a $1\times1$ Conv. 
The model was trained using the AdamW optimizer~\cite{adamw} (learning rate: $10^{-3}$, weight decay: $10^{-4}$). The batch size and loss weights are presented in {\cref{tab:loss_weights}}. Experiments were conducted using PyTorch~\cite{pytorch} on an NVIDIA RTX 4090 GPU.

\subsubsection{Compared Methods and Evaluation Metrics}

To demonstrate the superiority of our unified framework in both source-accessible and source-free adaptation, we compared it against state-of-the-art methods specifically designed for each respective setting. The baselines cover a variety of strategies, such as adversarial learning, variational inference, pseudo labels, image translation, distillation, contrastive learning, etc. Additionally, a source-trained attention U-Net (Att-UNet) was directly tested on the target domain, serving as a baseline without adaptation (w/o Adapt.). All methods were trained using the same data preprocessing pipeline to ensure a fair comparison. Evaluation metrics include Dice Similarity Coefficient (DSC) and Average Symmetric Surface Distance (ASSD).

\subsection{Comparison with State-of-the-Art Methods}

\cref{tab:result_sa_sf_mscmr,tab:result_sa_sf_amos} report quantitative results compared with state-of-the-art methods developed specifically for each respective setting. Notably,
our approach adopts a unified architecture under both settings, with the only difference being that, in the source-free case, the loss terms are split into two groups and applied in separate stages. Moreover, our method's adaptability emerges purely from the framework design, while all baseline methods heavily rely on \rev{R3.2}{explicit alignment objectives}{handcrafted adaptation strategies}, such as feature \rev{}{matching}{alignment} or pseudo-labeling\rev{}{}{ in the two settings, respectively}.

Across both datasets and settings, our method achieves the best performance in average DSC and ASSD and consistently outperforms prior approaches. The superiority is particularly pronounced under the source-free setting, where our method not only outperforms existing approaches by substantial margins across all evaluation metrics, but also narrows the long-standing performance gap with source-accessible models. Particularly on the {\mscmr} dataset, our model in the source-free setting even outperforms the best source-accessible baseline. On the {\amos} dataset, the Att-UNet without adaptation produces a very poor performance, indicating severe domain shift. Many source-free baselines fail because their adaptation strategies heavily rely on the initial predictions of the source-trained models, which are severely degraded. In contrast, the source-free variant of our method maintains a strong performance approaching that of the source-accessible counterpart. 
We attribute this improvement to our framework's ability to retain a semantically structured shape memory through learnable latent bases, leading to stronger generalizability and adaptability, even without persistent source supervision.

\begin{figure}[t]
  \centering
  \includegraphics[width=\linewidth]{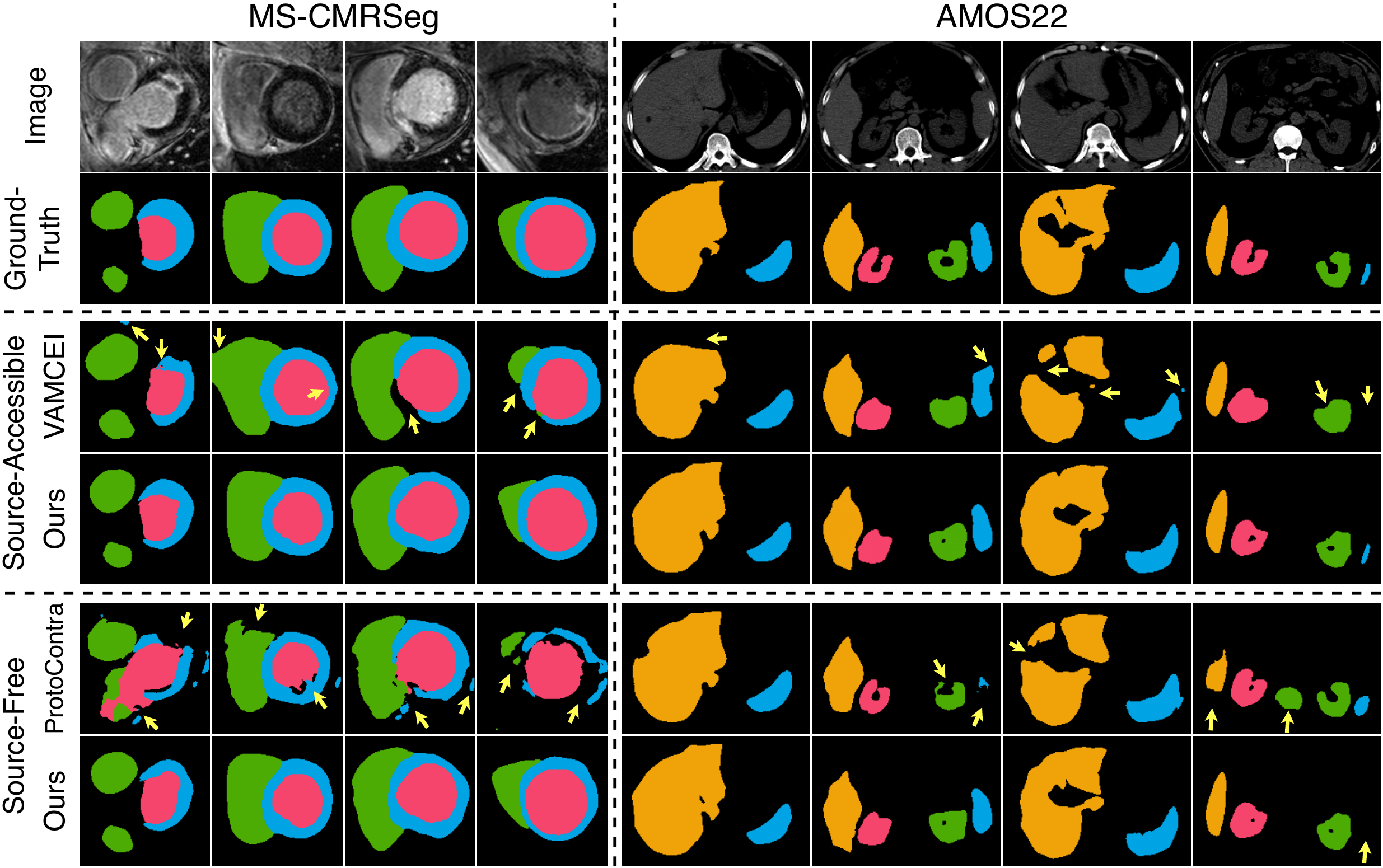}
  \caption{Qualitative comparison of our method and the baselines that achieve  best overall performance (VAMCEI/ProtoContra for the source-accessible/source-free settings). Yellow arrows indicate inferior results.
  }
  \label{fig:qualitative_compare}
\end{figure}

Qualitative comparisons in~\cref{fig:qualitative_compare} further highlight the strength of our approach. For challenging cases with poor image quality, low contrast, or imaging artifacts, even the best baselines produce physiologically invalid segmentations with fragmented shapes, particularly in the source-free setting. Our method effectively mitigates this issue in both settings, with predictions that preserve coherent topological structure. This robustness stems from the clear disentanglement of canonical anatomy and spatial deformation, which together ensure geometrically smooth and anatomically plausible predictions, even when pixel intensities are unreliable.

In summary, despite using a single model design, our method consistently outperforms all competing approaches, demonstrating superior generalizability, stability, and adaptability across multiple datasets and settings.

\subsection{Interpretability of Latent Manifold}
\label{sec:interpret}

While latent representations often contain entangled structures in conventional UDA studies, our model explicitly disentangles anatomical structure and individual-specific geometry through a low-dimensional and semantically organized latent manifold. In this section, we present a series of visualizations to illustrate the interpretability, generalizability, robustness and domain consistency of the proposed framework.

\subsubsection{Disentanglement of Canonical Anatomy and Geometry}

\begin{figure}[t]
  \centering
  \includegraphics[width=\linewidth]{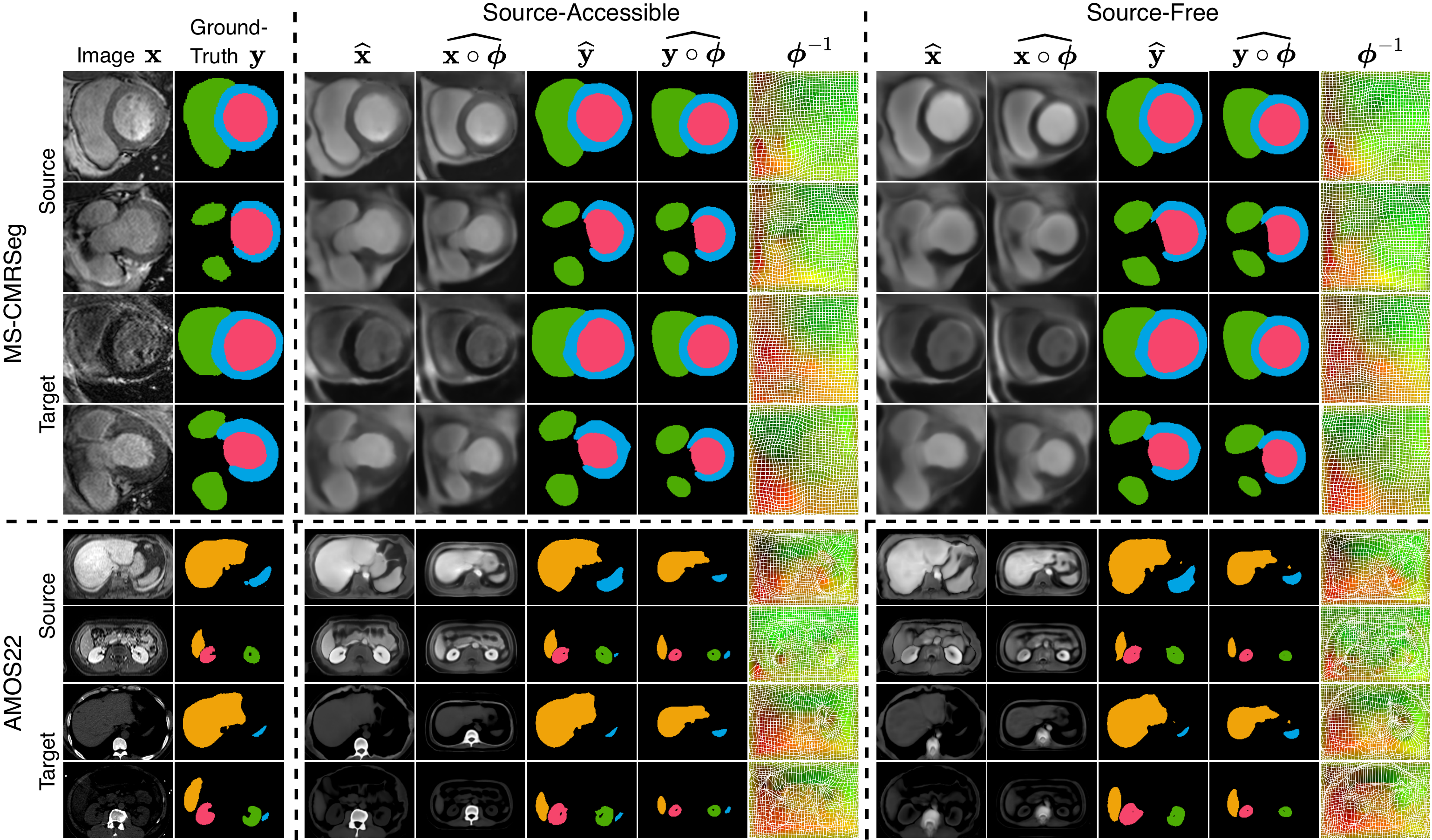}
  \caption{Disentanglement of canonical anatomy and geometry by our model. We visualize the templates $\z$ by decoding them into intermediate segmentations 
  $\widehat{\y\circ\st}$ and reconstructions $\widehat{\x\circ\st}$ using the segmentation and reconstruction decoders. We also show the corresponding deformations $\st^{-1}$, as well as the final segmentations $\widehat{\y}$ and reconstructions $\widehat{\x}$ obtained after warping by $\st^{-1}$.
  }
  \label{fig:template_deform}
\end{figure}

\cref{fig:template_deform} visualizes the disentanglement of canonical anatomy and individual-specific geometry for representative images. The results indicate that the templates $\z$ accurately capture the underlying anatomy topology and are semantically coherent across domains, illustrating that the learned latent manifold encodes domain-invariant structural priors. The deformations further adapt these templates to capture geometric variations, such as thickening, asymmetry, or tissue movement. This decoupling leads to both strong generalizability and anatomically plausible predictions.  

\subsubsection{Traversal of the Latent Composition Space}
To investigate the semantic structure of the latent manifold, we perform two types of traversals over the composition weights $\w$ via an interpolation operator $\mathcal{T}$, which manipulates the template $\z$. To respect the Riemannian geometry of the simplex $\Delta$ endowed by the Fisher-Rao metric, $\mathcal{T}$ proceeds along the geodesics on the positive orthant $\mathbb{S}_+^{M-1}$ of a unit sphere, \ie, $\mathcal{T}_\alpha(\w,\w^\prime)\defas ( [ \frac{\sin((1 - \alpha)\theta)}{\sin\theta} \sqrt{w_i} + \frac{\sin(\alpha \theta)}{\sin\theta} \sqrt{w^\prime_i} ]^2 )_{i=1}^M $, where $\alpha\in[0,1], \theta = \arccos( \sum_{i=1}^M \sqrt{w_i w^\prime_i})$.

\begin{figure}[t]
  \centering
  \includegraphics[width=\linewidth]{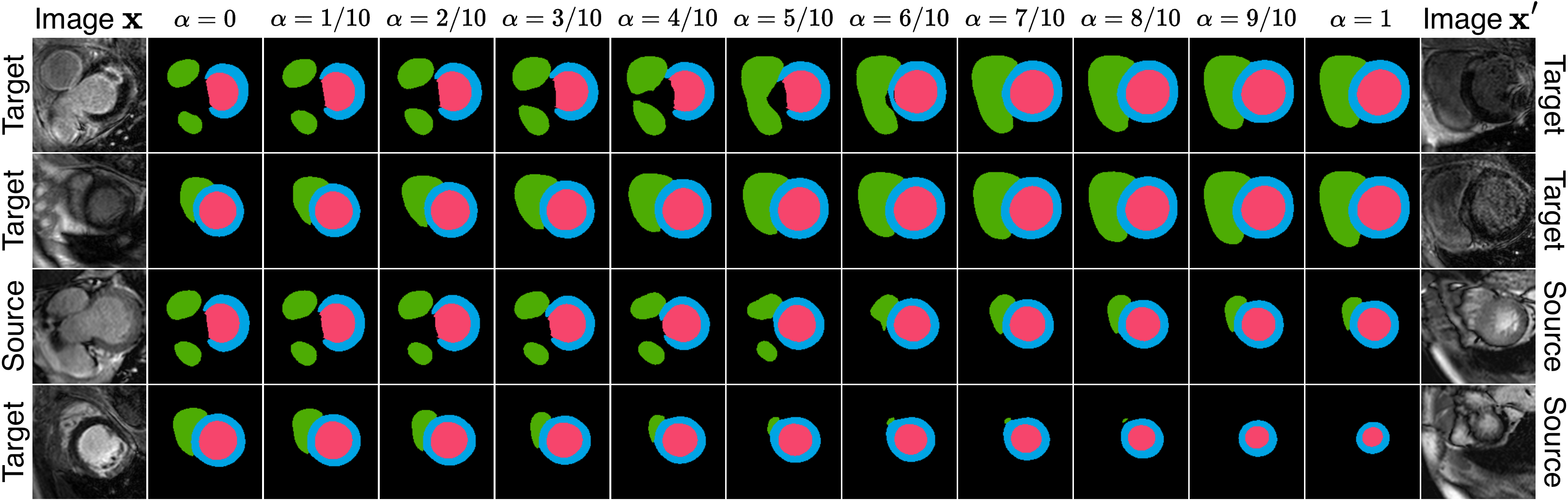}
  \caption{Inter-image traversal on the {\mscmr} dataset. Each row denotes the decoded segmentations corresponding to an interpolation $\mathcal{T}_\alpha(\w,\w^\prime)$ between the composition weights $\w,\w^\prime$ extracted from two images $\x,\x^\prime$. ``Target'' and ``Source'' indicates the image domains.
  }
  \label{fig:inter-image_traversal}
\end{figure}

\textbf{Inter-Image Traversal}: Given two images with distinct anatomical characteristics, we interpolate between their corresponding composition weights to generate $\z=\{\mubold^l\}_{l=1}^L$ through \cref{eq:qz_calculation} and decode them into segmentations. As shown in \cref{fig:inter-image_traversal}, the resulting templates transition smoothly and plausibly, reflecting a continuous shape morphing. This confirms that the latent space captures semantically meaningful variations. Notably, interpolation between source and target-domain samples yields similarly coherent transitions, underscoring domain-invariance of the learned manifold. 

\begin{figure}[t]
  \centering
  \includegraphics[width=\linewidth]{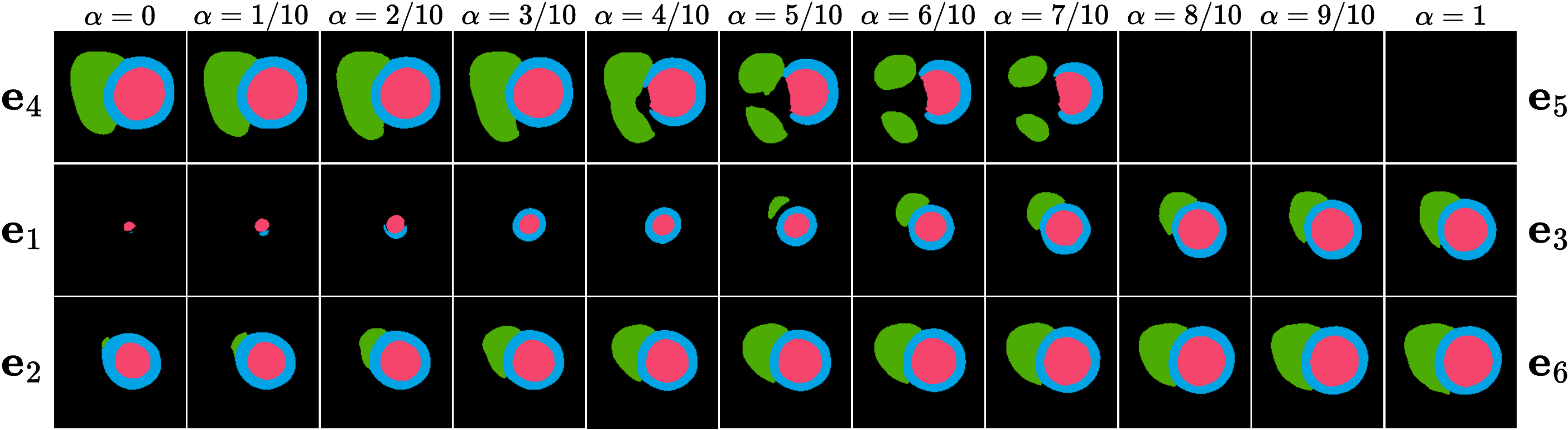}
  \caption{Inter-basis traversal on the  {\mscmr} dataset. Each row denotes the decoded segmentations corresponding to an interpolation $\mathcal{T}_\alpha(\e_i,\e_j)$ between a pair of one-hot composition weights $\e_i,\e_j$. All topological patterns observed in the displayed segmentations are anatomically valid, as some ground-truth labels in the dataset exhibit the same structures.
  }
  \label{fig:inter-basis_traversal}
\end{figure}

\textbf{Inter-Basis Traversal}: We further explore the semantic content encoded in the basis distributions. Specifically, we construct two one-hot vectors $\e_i,\e_j\in\mathbb{R}^M$, \ie, the composition weights that select only the base distributions $\{q_i(\z^l)\}_l,\{q_j(\z^l)\}_l$, respectively, and interpolate between them as $\mathcal{T}_\alpha(\e_i,\e_j)$, which are decoded into segmentations (\cref{fig:inter-basis_traversal}), similar to inter-image traversal. 
The results reveal smooth transitions, indicating that the learned bases form a diverse set of morphological primitives, which can be meaningfully blended via simplex-weighted mixture. 
In addition to interpretability, this traversal offers a diagnostic tool: if interpolations involving certain bases yield incoherent shapes or lack variability, those bases may be underutilized or redundant. We observe that the loss $\Lbalance$ for basis usage plays a key role in maintaining basis diversity and preventing mode collapse. 

Together, these two forms of traversal demonstrate that our model learns an anatomical plausible and geometrically smooth latent space. The ability to maintain structural coherence when moving continuously through this space enables both emergent adaptability and interpretation of the latent anatomy, as well as potential applications such as data augmentation, anomaly characterization, and interactive editing.

\subsubsection{Cross-Domain Alignment}

\begin{figure}[t]
  \centering
  \includegraphics[width=\linewidth]{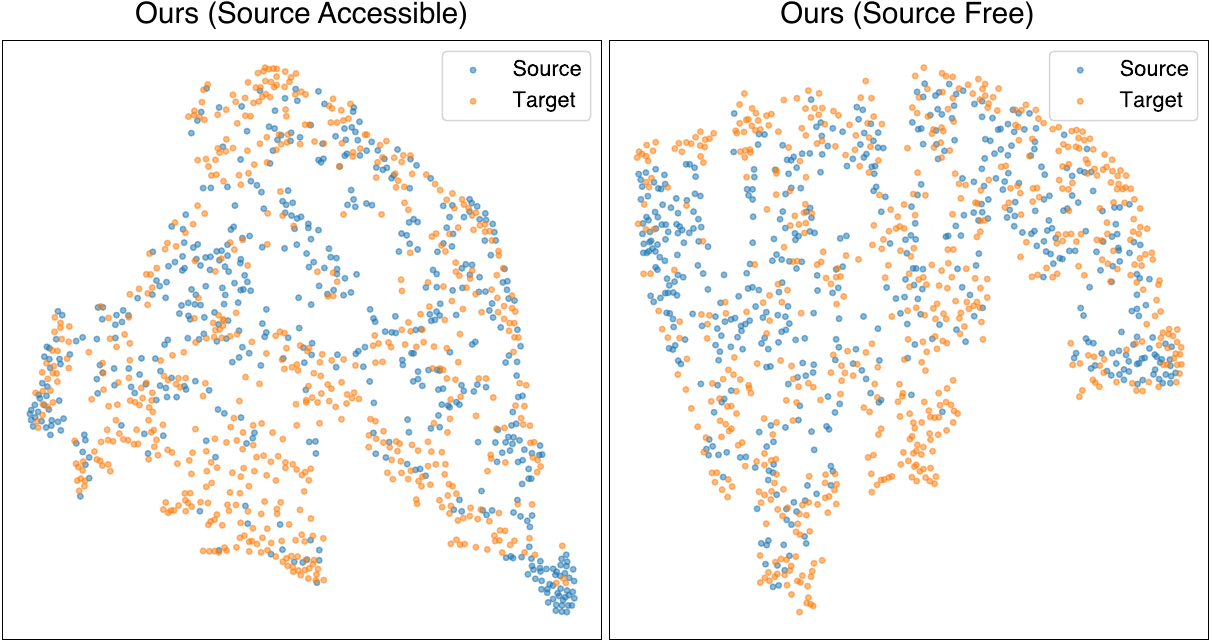}
  \caption{t-SNE results by our method on the {\mscmr} dataset in the source-accessible and source-free settings. The projection maps 6D composition weights to a 2D space.
  }
  \label{fig:tsne}
\end{figure}

To visualize how our approach harmonizes the latent spaces of source and target domains, we project the predicted composition vectors $\w$ from the two domains onto a 2D space using t-SNE~\cite{tSNE}. Notably, since $\w$ are low-dimensional (6D for the {\mscmr} dataset), t-SNE introduces minimal information loss. \cref{fig:tsne} shows the results in both settings, where the source and target samples are well aligned and form overlapping clusters. This demonstrates that our approach effectively achieves domain alignment. To the best of our knowledge, this is the first work that \rev{R3.2}{harmonizes source and target representations without explicit cross-domain alignment objectives}{achieves domain alignment without any explicit adaptation strategies} in both source-accessible and source-free settings.

\begin{table}[t]
\caption{Ablation studies on {\mscmr} by setting corresponding loss weights to 0 \rev{}{(for w/o $\Lznew$)}{} or a large value (for w/o $\st$).}
\label{tab:ablation_loss_weight}
\centering
\footnotesize
\setlength{\tabcolsep}{4pt}
\begin{tabular}{cccc}
\toprule
Method & DSC (\%) $\uparrow$ &ASSD (mm) $\downarrow$ & \makecell{Epochs to\\Converge} \\ \midrule
w/o $\boldsymbol{\phi}$ & 58.4$\pm$7.59   & 5.88$\pm$1.07  & 24       \\ 
w/o $\Lznew$ &82.1$\pm$6.02 & 1.96$\pm$0.78 & 2018\\
Proposed &84.1$\pm$5.35 & 1.72$\pm$0.80 & 2067 
\\\bottomrule
\end{tabular}
\end{table}

% \begin{table}[t]
% \caption{Ablation studies for the number of bases $M$ on the {\mscmr} dataset.}
% \label{tab:basis_ablation}
% \centering
% \footnotesize
% \setlength{\tabcolsep}{2.5pt}
% \begin{tabular}{cccccc}
% \toprule
% $M$ & 3 & 4 & 6 & 8 & 10 \\ \midrule
% DSC (\%) & 80.8$\pm$13.9 & 81.3$\pm$5.94 & 84.1$\pm$5.35 & 81.2$\pm$9.27 & 80.4$\pm$9.45 \\
% ASSD (mm) & 2.28$\pm$2.60 & 2.12$\pm$0.78  &1.72$\pm$0.80& 2.12$\pm$1.19 & 2.22$\pm$1.16
% \\\bottomrule
% \end{tabular}
% \end{table}

\subsection{Ablation Studies}

We conduct comprehensive ablation studies
\rev{}{}{in the source-accessible setting} to assess the effectiveness and necessity of each component in our framework. \rev{}{}{Most ablations are implemented by manipulating individual loss term weights, as summarized in Table V.}

\begin{table}[t]
    \centering
    \footnotesize
    \setlength{\tabcolsep}{1.5pt}
    \caption{\rev{}{Quantitative comparison of segmentation and reconstruction performance between our disentanglment architecture and a direct-decoding baseline on the source domain of the two datasets. ASSD is reported in millimeters (mm), and PSNR is reported in decibels (dB).}{}}
    \begin{tabular}{ccccccc}
    \toprule
      \multirow{2.5}{*}{Dataset} & \multirow{2.5}{*}{Model}   & \multicolumn{2}{c}{Segmentation} && \multicolumn{2}{c}{Reconstruction}{} \\
      \cmidrule{3-4} \cmidrule{6-7}
     & &DSC (\%) $\uparrow$ & ASSD $\downarrow$ & & PSNR $\uparrow$& SSIM $\uparrow$\\
      \midrule
   \multirow{2}{*}{\mscmr} &  Direct-Dec  & 92.4$\pm$1.08 & 0.41$\pm$0.13 & & 35.4$\pm$4.08 & 0.982$\pm$0.007\\
&   Ours-Sup    & 92.7$\pm$0.99 & 0.38$\pm$0.17 & & 20.6$\pm$1.13 & 0.611$\pm$0.026\\

      \midrule
   \multirow{2}{*}{\amos} &  Direct-Dec  & 94.7$\pm$2.13 & 1.17$\pm$0.39 && 36.8$\pm$0.65 & 0.966$\pm$0.019 \\
&   Ours-Sup    & 94.5$\pm$1.21 & 1.44$\pm$0.14 & & 21.1$\pm$0.85 & 0.737$\pm$0.034 \\
      \bottomrule
    \end{tabular}
    
    \label{tab:direct-decoding}
\end{table}

\subsubsection{\rev{R2.1}{Pixel-Wise Expressiveness for Dense Prediction}{}}
\label{spatial_bottleneck_ablation}
\rev{}{
This ablation evaluates whether the proposed canonical-deformation factorization
limits pixel-wise spatial prediction.
To this end, we compare two models under a fully supervised setting:
(i) {Ours-Supervised}, which uses our full architecture to model image structure; and
(ii) {Direct-Decoding}, where segmentation and reconstruction are predicted
directly from dense image features using two decoders attached to the same encoder.
Both models use the same encoder and decoder backbones and are trained with the same
segmentation and reconstruction losses.
As shown in \cref{tab:direct-decoding}, our model achieves segmentation accuracy
comparable to direct decoding on both datasets, including the more complex
multi-organ scenario.
This result indicates that relying on a low-dimensional blending vector and
deformation-based warping does not impose a spatial-bandwidth bottleneck for
pixel-wise segmentation, nor increase learning difficulty in practice.
Moreover, maintaining segmentation accuracy while sharing the bases across
reconstruction, segmentation, and deformation prediction suggests that the
global bases are not overburdened by multi-task usage, but instead capture an
anatomical substrate that is jointly useful across tasks.
While direct decoding attains higher reconstruction fidelity, this difference
is expected due to its unconstrained use of dense features and does not
contradict the strong segmentation accuracy achieved by our method.
}{}

\subsubsection{Anatomical Disentanglement}

\rev{}{This ablation evaluates}{ We first verify} the effectiveness of disentangling anatomy and geometry by evaluating segmentation performance without spatial transformation. Specifically, by setting the loss weight $\lambda_3$ to a large constant, the inferred deformation field collapses to identity. As a result, segmentation is directly decoded from the canonical anatomical template without warping.
As shown in \cref{tab:ablation_loss_weight} row 2, removing deformation leads to a noticeable drop in segmentation accuracy. This confirms that modeling image-specific geometric variations is critical for adapting canonical shape priors to individual anatomy, and the disentanglement of anatomy and geometry leads to greatly improved accuracy, which aligns with human visual recognition.

\rev{R1.1, \\R4.6}{}{
2) Basis Expressiveness: To investigate the expressiveness of the learned basis distributions, we conduct two complementary ablations: 
a) Removing the basis usage loss $\Lbalance$ (
% \cref{tab:ablation_loss_weight}
Table IV
row 3), which increases the risk of mode collapse where fewer bases are activated. b) Varying the number of bases $M$. As shown in {
% \cref{tab:basis_ablation}
Table V
}, reducing $M$ leads to degraded performance since it constrains the optimization space and limits the diversity of anatomical templates, while overly large $M$ introduces redundancy and optimization instability.}

\subsubsection{\rev{R3.3}{Effect of Structural and Usage Regularizers}{}}

\rev{}{This ablation analyzes the effects of $\Lwdice$ and $\Lbalance$.
As these regularizers are employed in both source-accessible and source-free
settings, we conduct the analysis under the source-free setting, where the
two-stage training procedure enables a more comprehensive, stage-wise
examination of their roles. }{}
\paragraph{\rev{}{Experimental Design}{}}
\rev{}{In Stage-1 (source-supervised training), we consider four model variants,
denoted as S1--S4, corresponding to all combinations of enabling or disabling $\Lwdice$ and $\Lbalance$.
Specifically, S1 uses neither regularizer, S2 uses $\Lwdice$ only, S3 uses
$\Lbalance$ only, and S4 uses both regularizers.
This $2\times2$ design allows us to isolate the individual and }{}\rev{}{combined
effects of the two regularizers on the learned latent representation. In Stage-2 (target-only adaptation), $\Lwdice$ is disabled by design due to the absence of target-domain labels, and thus we
compare two variants (both initialized from
the full Stage-1 configuration S4): T1, where $\Lbalance$ is removed,
and T2 (ours), where $\Lbalance$ is retained.
This comparison isolates the marginal effect of $\Lbalance$ during
unsupervised adaptation under a fixed latent manifold.}{}

\paragraph{\rev{}{Evaluation Metrics}{}}
\rev{}{To quantitatively characterize how $\Lwdice$ and $\Lbalance$
affect the learned latent variables, we report both segmentation
accuracy and a set of diagnostic metrics:}{}
\begin{itemize}
    \item \rev{}{\textbf{Basis utilization.}
Let $\{\w_i\in \Delta^{M-1}\}_{i=1}^N$ denote the blending vectors
predicted for a dataset of $N$ images.
We define the dataset-level basis usage rate as
$
\overline{\w} = \frac{1}{N}\sum_{i=1}^N \w_i ,
$
where the $m$-th element $\overline{w}_m$ represents the average usage of
the $m$-th basis.
To measure how evenly the bases are utilized, we compute the usage entropy
$
H(\overline{\w}) = -\sum_{m=1}^M \overline{w}_m \log \overline{w}_m ,
$
and the number of effective bases
$
N_{\mathrm{eff}} = \exp\!\big(H(\overline{\w})\big),
$
which reaches its maximum value $M$ when all bases are used uniformly.}{}
\item \rev{}{\textbf{Simplex geometry.} To quantify how broadly the blending vectors occupy the simplex
$\Delta^{M-1}$, we define a dispersion metric based on the generalized
variance.
Since $\Delta^{M-1}$ lies in a $(M-1)$-dimensional affine subspace of
$\mathbb{R}^M$, we first center the samples by
$\widetilde{\w}_i = \w_i - \overline{\w}$ and project them onto the intrinsic
subspace $\{\boldsymbol{u} \in \mathbb{R}^M : \mathbf{1}^\top \boldsymbol{u} = 0\}$ using a fixed
orthonormal basis $U \in \mathbb{R}^{M \times (M-1)}$.
The empirical covariance is computed as $\Sigma = \frac{1}{N-1}\sum_{i=1}^N (U^\top \widetilde{\w}_i)
(U^\top \widetilde{\w}_i)^\top ,$
and the dispersion is then measured by
$
Q = \log \det(\Sigma + \epsilon I),$
with $\epsilon=10^{-6}$ for numerical stability.
Larger values of $Q$ indicate a more dispersed distribution of blending
vectors on the simplex.}{}
\item \rev{}{\textbf{Structural consistency.}
We also compute the Spearman rank correlation $r_s$
between blending vector distances $\fr{\w_i}{\w_j}$ and canonical segmentation dissimilarities $
1 - \mathrm{DSC}\big(\y_i \circ \st_i,\; \y_j \circ \st_j\big)
$ over all sample pairs $(i,j)$.
A high value of $r_s$ indicates that variations in $\w$ consistently induce
corresponding changes in the segmentation outcomes.}{}
\end{itemize}

\begin{figure}[t]
    \centering
    \includegraphics[width=\linewidth]{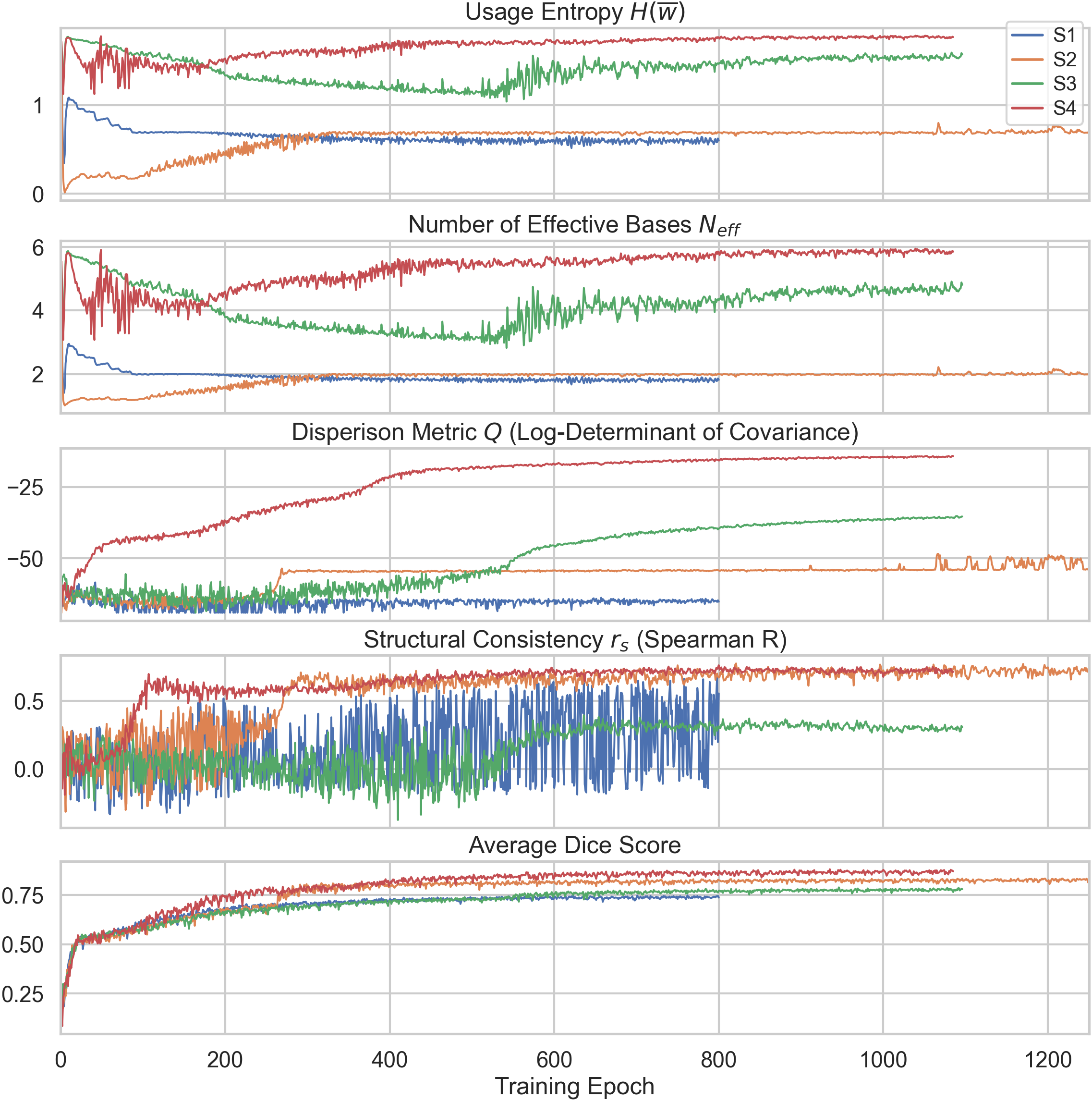}
    \caption{\rev{}{Stage-1 training dynamics and latent-space quality on the source domain under different regularizer configurations, including evolution of (a) usage entropy $H(\overline{\w})$, (b) number of effective bases $N_{\mathrm{eff}}$, (c) dispersion metric $Q$, (d) structural consistency $r_s$, and (e) segmentation Dice during source-domain supervised training. All curves share the same training timeline.}{}}
    \label{fig:stage1_dynamics}
\end{figure}

\begin{figure}[t]
    \centering
    \includegraphics[width=\linewidth]{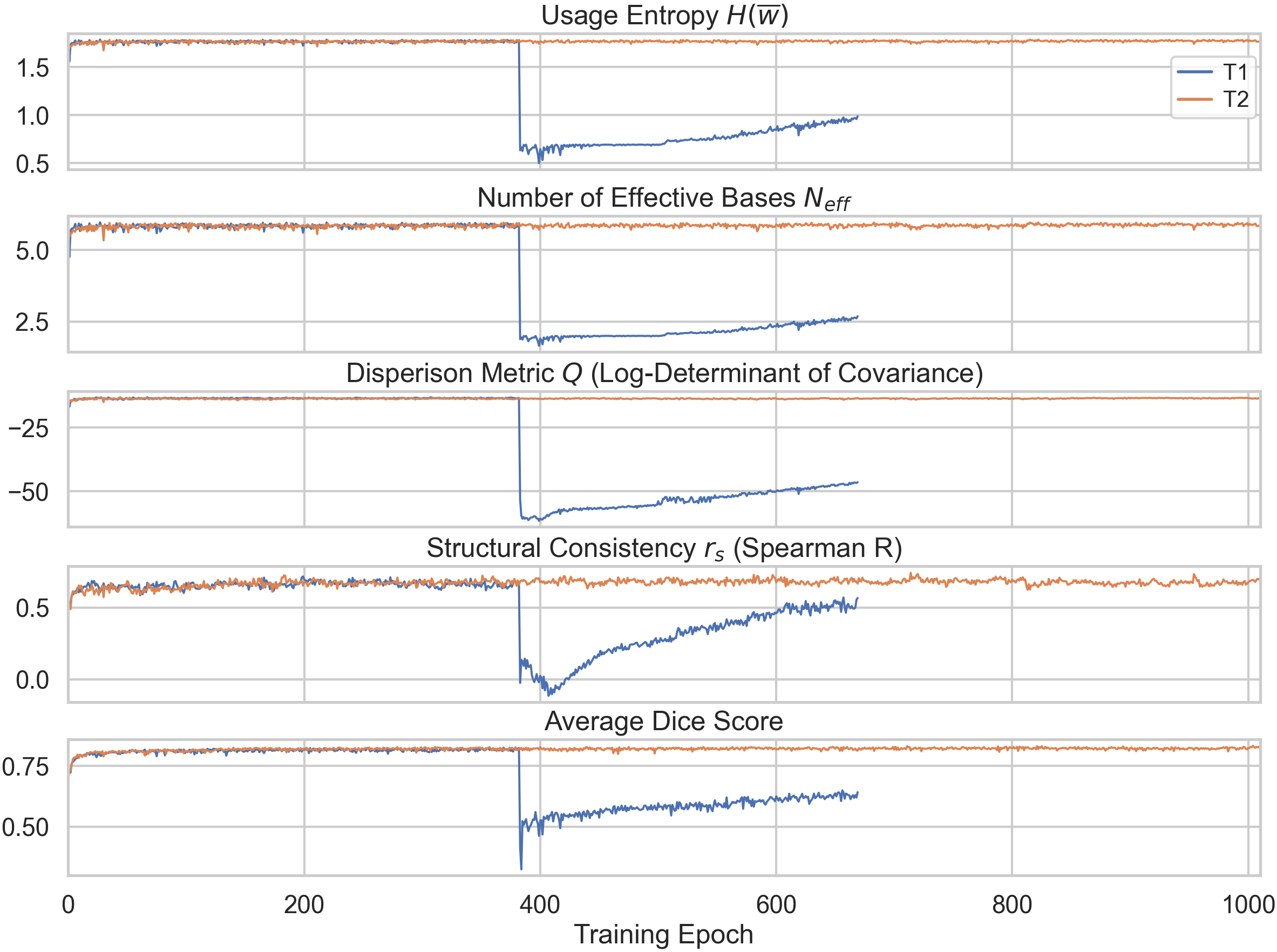}
    \caption{\rev{}{Stage-2 adaptation dynamics on the target domain, including evolution of basis-utilization metrics, simplex-geometry metrics, and target-domain segmentation Dice during unsupervised adaptation, comparing models without ($T1$) and with ($T2$) the usage regularizer. The latent simplex learned in stage~1 is kept fixed.}{}}
    \label{fig:stage2_dynamics}
\end{figure}

\paragraph{\rev{}{Results}{}}

\rev{}{\cref{fig:stage1_dynamics} summarizes the Stage-1 training dynamics under different regularizer
configurations (S1--S4).
Without $\Lbalance$, basis utilization remains highly uneven throughout training,
with persistently low usage entropy and $N_{\mathrm{eff}}$ (S1, S2).
Enabling $\Lbalance$ substantially increases both metrics (S3, S4),
indicating broader and more stable basis participation.
In contrast, $\Lwdice$ primarily affects the geometric organization of the latent space:
settings with $\Lwdice$ (S2, S4) exhibit markedly higher dispersion $Q$, when compared with variants under the same usage-regularizer setting,
as well as stronger structural consistency $r_s$.
The full model (S4) consistently achieves the highest values across all metrics,
corresponding to faster convergence and higher segmentation accuracy.
\cref{fig:stage2_dynamics} shows the Stage-2 adaptation dynamics on the target domain.
When $\Lbalance$ is removed during adaptation (T1),
basis utilization exhibits an abrupt drop,
accompanied by a collapse in dispersion $Q$ and a decrease in $r_s$.
This degradation coincides with unstable target-domain Dice.
In contrast, retaining $\Lbalance$ (T2) preserves balanced basis utilization
and stable simplex geometry, leading to smooth and monotonic improvement
in segmentation performance.
These results indicate that while 
the latent manifold is fixed during Stage-2,
the usage regularizer plays a critical role in stabilizing how target-domain
samples populate and exploit the learned simplex.}{}

\begin{figure}[t]
    \centering
    \includegraphics[width=\linewidth]{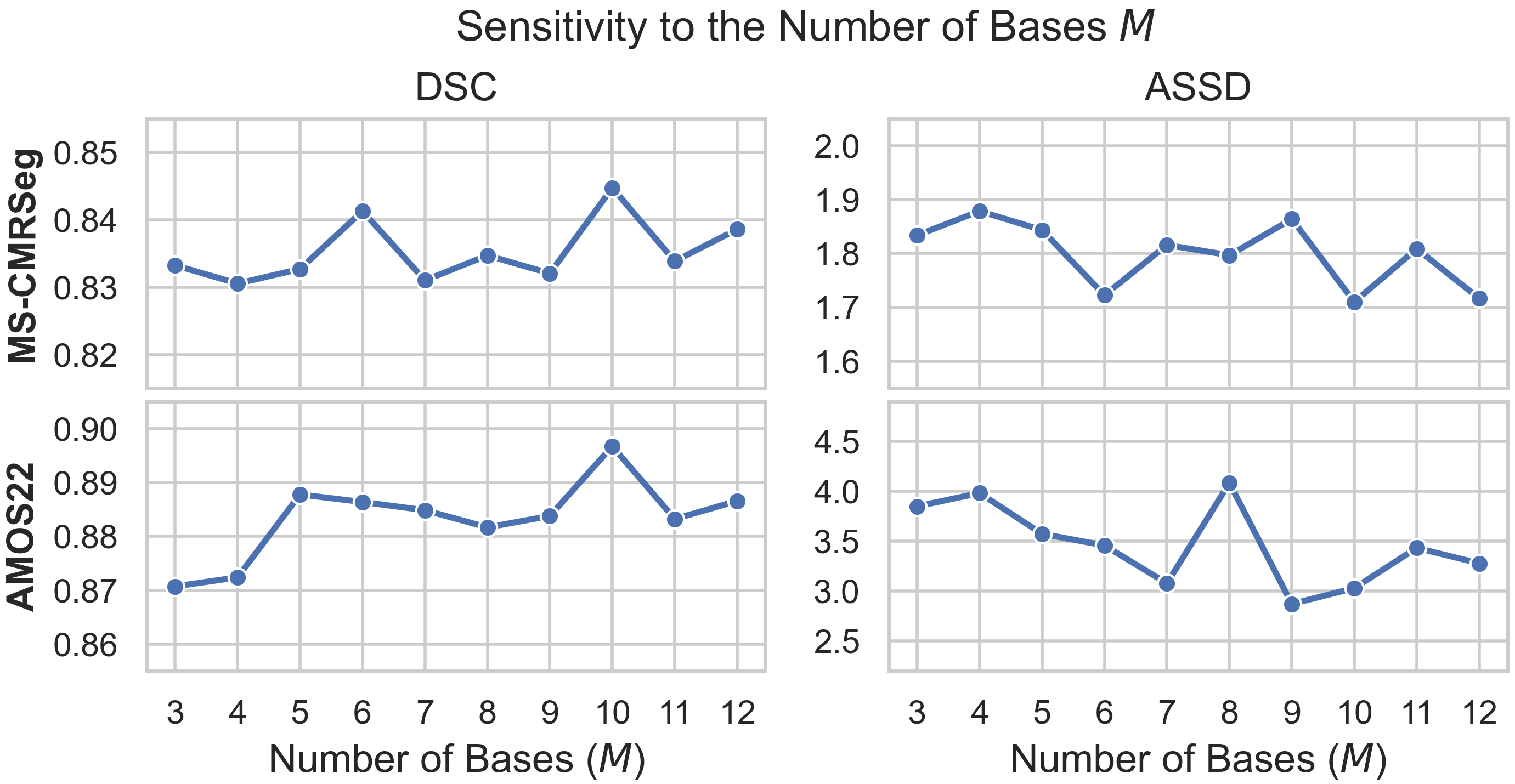}
    \caption{\rev{}{Sensitivity analysis with respect to the number of bases $M$.
Segmentation performance is evaluated using DSC (left column) and ASSD (right column, in mm) on {\mscmr} (top row) and {\amos} (bottom row).}{}}
    \label{fig:best_M}
\end{figure}

\subsubsection{\rev{}{Robustness to the Number of Bases}{}}
\rev{R1.1,\\R4.5}{
This ablation examines the effect of the number of bases $M$ by sweeping $M$ from 3 to 12 on the
two datasets, while keeping all other training and evaluation settings fixed.
As shown in \cref{fig:best_M}, performance remains consistently strong over a wide
range}{} \rev{}{of $M$ values on both datasets, forming a broad performance plateau.
Importantly, neither dataset exhibits a preference for a narrowly tuned $M$.
For the more anatomically complex {\amos} dataset, slightly increased sensitivity is
observed only at very small $M$, while performance quickly stabilizes as $M$
increases.
Overall, these results demonstrate that our method is robust to the choice of $M$. A
moderately sized basis set is sufficient in practice, without
requiring detailed tuning.
}{}

\begin{table}[t]
    \centering
    \footnotesize
    \setlength{\tabcolsep}{2pt}
    \caption{\rev{}{Target-domain segmentation and reconstruction performance of our model on the {\mscmr} dataset after Stage-1 training and after Stage-2 target-only adaptation.
Stage-1 refers to supervised training on the source domain, and Stage-2 refers to subsequent adaptation using unlabeled target data. $\Delta$ denotes the performance change from Stage-1 to Stage-2.}{}}
    \begin{tabular}{ccccccc}
    \toprule
     \multirow{2.5}{*}{Training Stage}   & \multicolumn{2}{c}{Segmentation} && \multicolumn{2}{c}{Reconstruction}{} \\
      \cmidrule{2-3} \cmidrule{5-6}
 &DSC (\%) $\uparrow$ & ASSD (mm) $\downarrow$ & & PSNR (dB) $\uparrow$& SSIM $\uparrow$\\
      \midrule
Stage 1  &  74.0$\pm$10.3 & 3.16$\pm$1.43 && 17.9$\pm$0.83 & 0.420$\pm$0.045\\
Stage 2    & 83.1$\pm$5.55 & 1.88$\pm$0.77 && 21.8$\pm$1.21 & 0.514$\pm$0.063\\

      \midrule
 $\Delta$ &  $+$9.1 & $-$1.28 && $+$3.9 & $+$0.094\\
      \bottomrule
    \end{tabular}
    \label{tab:stage2delta_mscmr}
\end{table}

\begin{table}[t]
    \centering
    \footnotesize
    \setlength{\tabcolsep}{2pt}
    \caption{\rev{}{Target-domain segmentation and reconstruction performance of our model on the {\amos} dataset after Stage-1 training and Stage-2 target-only adaptation.}{}}
    \begin{tabular}{ccccccc}
    \toprule
     \multirow{2.5}{*}{Training Stage}   & \multicolumn{2}{c}{Segmentation} && \multicolumn{2}{c}{Reconstruction}{} \\
      \cmidrule{2-3} \cmidrule{5-6}
 &DSC (\%) $\uparrow$ & ASSD (mm) $\downarrow$ & & PSNR (dB) $\uparrow$& SSIM $\uparrow$\\
      \midrule
Stage 1  & 48.9$\pm$17.7 & 20.8$\pm$7.94 && 12.5$\pm$1.13 & 0.303$\pm$0.050\\
Stage 2    &  87.0$\pm$3.27 & 3.28$\pm$1.30 && 17.5$\pm$0.27 & 0.570$\pm$0.044\\

      \midrule
 $\Delta$  &  $+$38.1 & $-$17.52 && $+$5.0 & $+$0.267\\
      \bottomrule
    \end{tabular}
    \label{tab:stage2delta_{\amos}}
\end{table}

\begin{figure}[t]
    \centering
    \includegraphics[width=\linewidth]{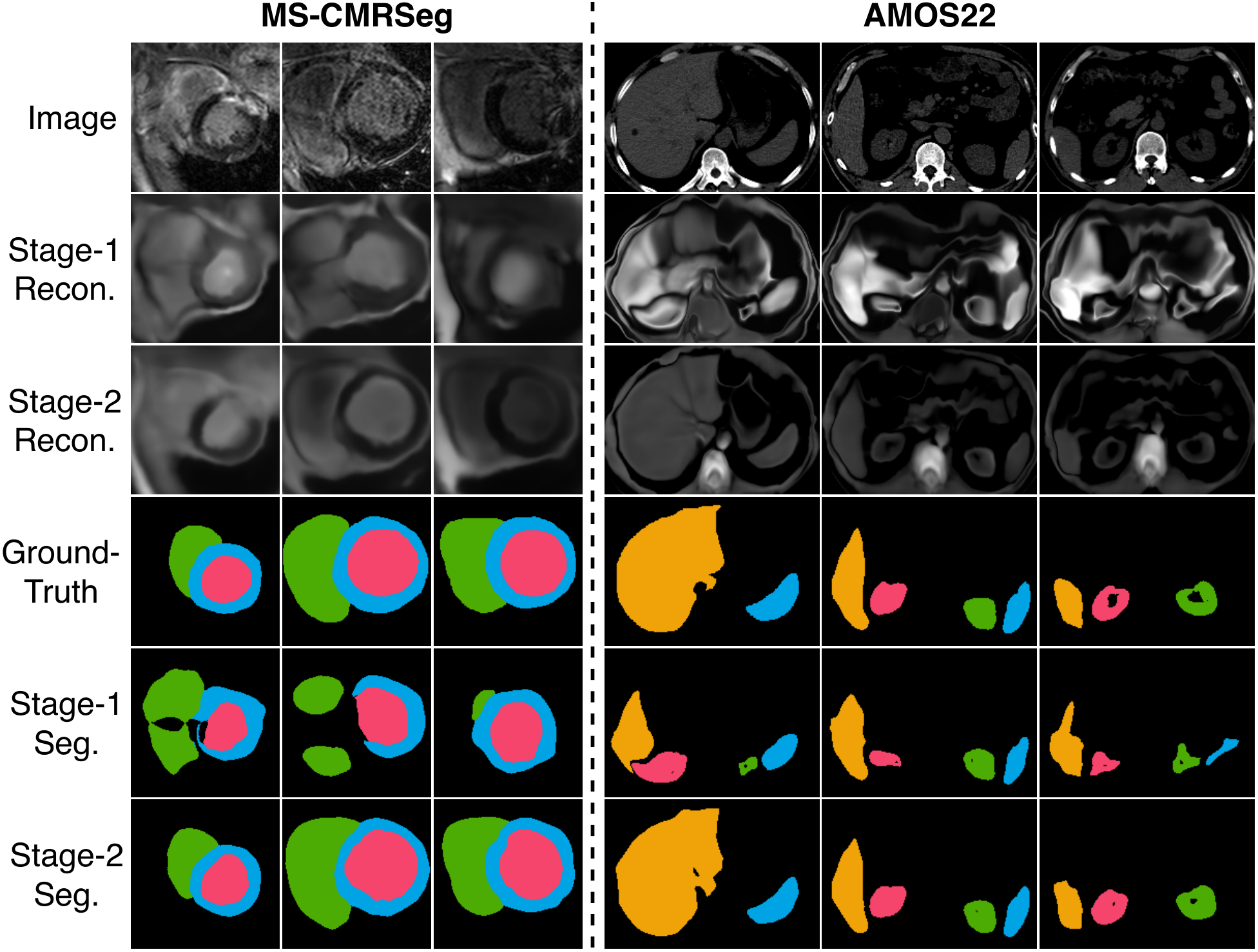}
    \caption{
\rev{}{Qualitative comparison of model outputs on representative target-domain cases after Stage-1 and Stage-2 training.}{}}
    \label{fig:two_stage_seg_recon_change}
\end{figure}

\subsubsection{\rev{}{Effect of Stage-2 in the Source-Free Setting}{}}
\rev{R2.4}{
This ablation investigates the contribution of Stage-2 target-only adaptation
beyond Stage-1 source-supervised training in the source-free setting. While Stage-1 learns semantic anatomical representations under source supervision,
Stage-2 recalibrates appearance-sensitive mappings to align unlabeled target images
with the learned semantic manifold.
As shown in \cref{tab:stage2delta_mscmr,tab:stage2delta_{\amos}}, directly applying the Stage-1 model to the target
domain leads to substantial performance degradation, while Stage-2 yields
consistent and non-trivial improvements across both datasets.
The paired visualizations in \cref{fig:two_stage_seg_recon_change} further show that Stage-2 systematically
corrects appearance-induced reconstruction biases and stabilizes anatomical
structures, which in turn leads to improved segmentation accuracy.
These results demonstrate that Stage-2 is not a minor fine-tuning step, but
plays a critical role in recalibrating target-domain image–manifold mappings
under unlabeled data.
}{}

\begin{figure}[t]
    \centering
    \includegraphics[width=\linewidth]{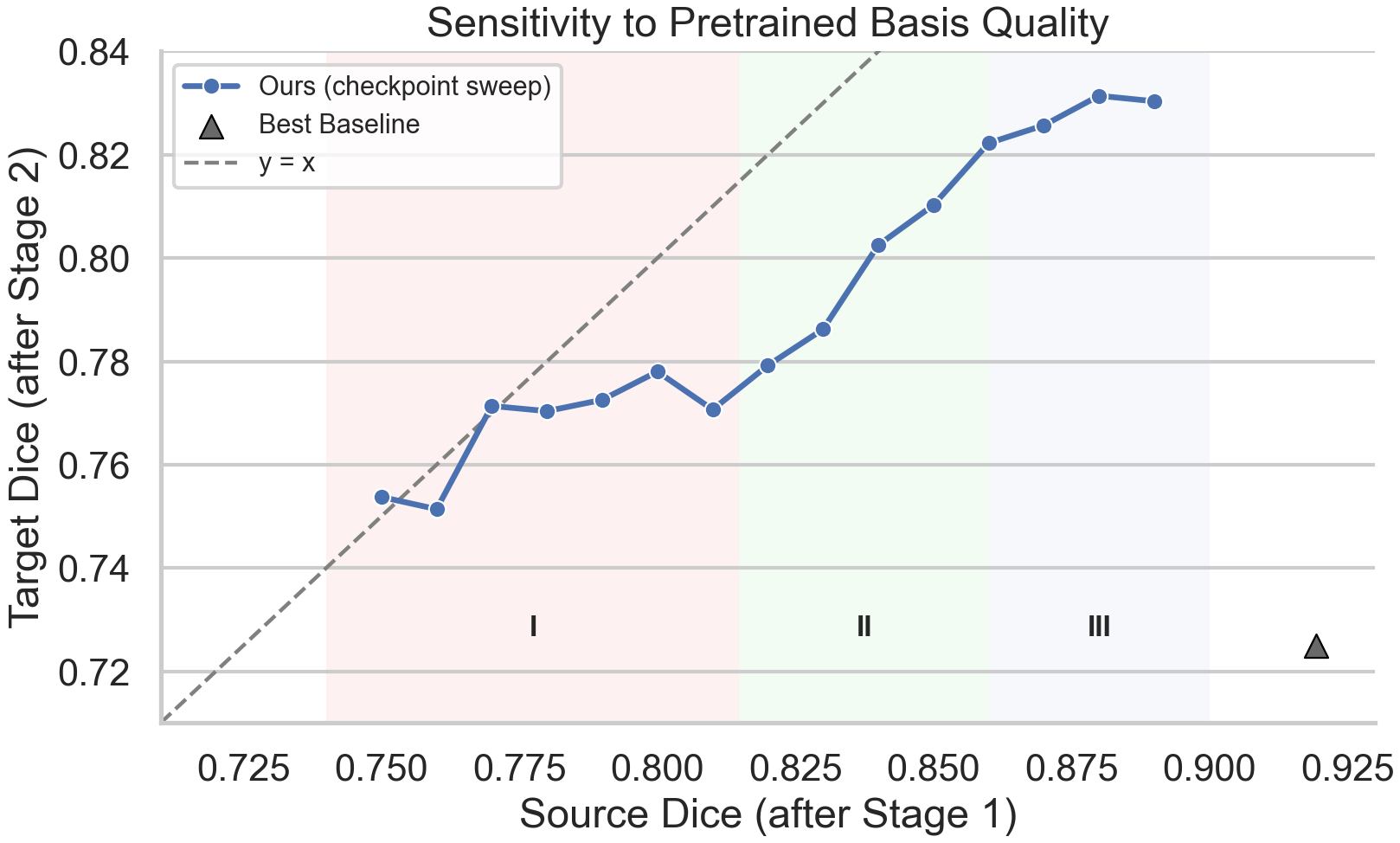}
    \caption{\rev{}{Sensitivity of source-free adaptation performance to the quality of
    source pretraining on MS-CMRSeg.
    Target-domain Dice after Stage-2 adaptation is shown as a function of the
    source-domain Dice achieved at the end of Stage-1.
    Each point corresponds to a pretrained checkpoint selected via early stopping.
    The dashed line $y=x$ indicates the theoretical upper bound.}{}}
    \label{fig:sensitivity_pretrained_basis_quality}
\end{figure}

\subsubsection{\rev{}{Sensitivity to Source Pretraining Quality in the Source-Free Setting}{}}
\rev{R4.4\\R4.9}{This ablation analyzes the sensitivity of the proposed source-free adaptation framework
to the quality of source pretraining by performing a checkpoint sweep on the
source-domain training stage.
Because the latent bases and segmentation decoder are fixed during Stage-2,
source pretraining quality defines the semantic prior available for adaptation.
Specifically, we early-stop source training at different validation Dice levels,
obtaining a series of pretrained models spanning from under-trained to near-optimal
initializations.
Each checkpoint is then used to initialize Stage-2 target-only adaptation under
identical settings.}{}

\rev{}{\cref{fig:sensitivity_pretrained_basis_quality} plots the target-domain Dice after
adaptation as a function of the source-domain Dice achieved at the end of Stage-1.
As expected, target performance depends on the reliability of source pretraining,
since it provides the only semantic prior in the source-free setting.
However, the empirical trend exhibits a much weaker dependence than the theoretical
upper bound $y=x$, indicating that target performance degrades gracefully as source
pretraining quality decreases.
We also observe three qualitative regimes: (i) under-trained source models lead to limited
adaptation gains, (ii) once a moderate source performance threshold is reached,
target adaptation becomes effective and stable, and (iii) further improvements in
source Dice yield diminishing returns on the target domain.
These results suggest that while source pretraining quality is necessary, the proposed
framework operates robustly across a}{} \rev{}{broad range of source initialization quality.}{}

\rev{}{For reference, we also report the performance of the strongest baseline
(ProtoContra) evaluated using its best-performing source-domain checkpoint.
Despite this favorable initialization, the baseline remains clearly below the
proposed method across the sensitivity range, including cases where our model
is initialized from substantially weaker source checkpoints.
This observation indicates that the superior target-domain performance of the
proposed framework cannot be attributed solely to stronger source pretraining,
but instead arises from its architectural design for source-free adaptation.}{}

\begin{table}[t]
\centering
\caption{\rev{}{Quantitative effect of removing the reconstruction loss $\Lx$ in the source-accessible setting.}{}}
\label{tab:recon_ablation}
\begin{tabular}{lcccc}
\toprule
Dataset & $\Lx$ & Dice (\%) $\uparrow$ & ASSD (mm) $\downarrow$ & \\
\midrule
\multirow{2}{*}{\mscmr} & \checkmark  & 84.1$\pm$5.35 & 1.72$\pm$0.80  \\
 & -- & 63.8$\pm$8.43 & 4.92$\pm$1.25  \\
\specialrule{0.3pt}{3pt}{3pt}
\multirow{2}{*}{\amos}    & \checkmark  & 89.7$\pm$1.30 & 3.03$\pm$1.12   \\
    & -- & 42.0$\pm$11.0 & 19.9$\pm$6.13   \\
\bottomrule
\end{tabular}
\end{table}

\subsubsection{\rev{}{Effect of the Reconstruction Loss}{}}
\rev{R4.6}{
This ablation evaluates the contribution of the reconstruction loss $\Lx$
by removing it from the training objective under the source-accessible setting,
while keeping all other components and hyperparameters unchanged.
As shown in \cref{tab:recon_ablation}, removing $\Lx$ leads to a substantial degradation in segmentation
accuracy on both datasets.
This behavior is consistent across datasets with different anatomical scope
and imaging characteristics.
These results indicate that the reconstruction pathway is essential for effective
adaptation.
Indeed, it provides the only image-based supervision that anchors the latent
manifold to the observed target images, and is also a necessary component implied
by the generative formulation as the observational likelihood.
}{}
\normalcolor 

\begin{table}[t]
\centering
% \small
\caption{\rev{}{Quantitative effect of hierarchical warping scales on {\mscmr}.
Different configurations correspond to using a subset $\Lambda$ of
multi-scale warping levels.}{}}
\label{tab:ablation_scales}
\setlength{\tabcolsep}{6pt}
\begin{tabular}{l c c}
\toprule
{Warping Scales $\Lambda$} 
& {DSC (\%) $\uparrow$} 
& {ASSD (mm) $\downarrow$} \\
\midrule
Single-scale ($\{5\}$) 
& 77.9$\pm$6.39 
& 2.59$\pm$0.87 \\

Reduced-scale ($\{3,5\}$) 
& 82.8$\pm$4.53 
& 1.90$\pm$0.69 \\

Ours ($\{1,3,5\}$) 
& {84.1$\pm$5.35} 
& {1.72$\pm$0.80} \\
\bottomrule
\end{tabular}
\end{table}

\subsubsection{\rev{R4.3}{Effect of Hierarchical Decomposition of Anatomical Structure}{}}
\label{ablation_hierarchical}

\rev{}{This ablation studies the effect of hierarchical anatomical priors by varying the set of
canonical-space warping scales used in the model.
In our formulation, anatomical bases are defined across a fixed levels $\{1,\dots,L\}$ of
network resolutions, while hierarchical composition is controlled by selecting
a subset $\Lambda$ of these resolutions at which deformation is applied, as detailed in \cref{eq:hierarchical_decomp}.
This subset governs how anatomical variation is
distributed across spatial scales.}{}

\rev{}{We compare three variants: $\Lambda=\{5\}$ (single full-resolution scale), $\{3, 5\}$ (reduced-scale), and $\{1,3,5\}$ (ours).
All variants share the same backbone, latent bases, and training protocol.
As summarized in \cref{tab:ablation_scales}, segmentation performance
improves consistently as additional warping scales are incorporated.
These results indicate that hierarchical anatomical priors, realized through
multi-scale warping, play an important role in capturing structural variability.
In principle, using a larger set of scales (e.g., $\Lambda=\{1,\dots,5\}$)
could further increase modeling capacity.
Accordingly, we adopt $\Lambda=\{1,3,5\}$ as a practical trade-off between
representational expressiveness and computational cost.}{}

\subsubsection{\rev{}{Effect of the Template Loss}{Manifold Structuring}}

\rev{}{This ablation evaluates the effect of $\Lznew$.}{
We also evaluate the remaining loss terms designed to better structure the latent manifold.
a) Removing $\Lwdice$, which encourages differences in $\w$ to reflect anatomical dissimilarity, resulted identical $\w$ across images during early training phases, and significantly impaired convergence speed, as well as degraded performance (Table V row 4). This validates the effectiveness of $\Lwdice$ in accelerating training, enhancing the diversity of $\w$ and structuring the manifold.
b) }
\rev{}{As shown in \cref{tab:ablation_loss_weight} row 3, r}{
R}emoving the KL $\Lznew$ over basis distributions causes the bases to drift independently, weakening segmentation performance\rev{}{}{ (Table V row 5)}. This highlights the value of basis regularization in preserving a coherent manipulation of the template $\z$ through $\w$ for effective adaptation.

\section{\rev{}{Limitations and Future Directions}{}}

\rev{}{Despite the strong empirical performance of the proposed framework, several limitations warrant discussion.}{}

\subsubsection{\rev{}{2.5D/3D Extensions}{}}

\rev{R1.1, \\R3.4}{Our method is in 2D, as the considered datasets exhibit substantial through-plane anisotropy. Under these conditions, enforcing a 3D or 2.5D architecture may introduce inappropriate inductive bias. Moreover, extending the method to 3D dramatically increases requirements on memory, training time, and annotated data to ensure stable optimization.
Nevertheless, the method remains straightforward to extend to 2.5D or 3D variants in implementation, such as slice-stacked encoders or volumetric canonical spaces with 3D deformations, when sufficiently isotropic data are available.}{}

\subsubsection{\rev{}{Cross-slice Consistency}{}}

\rev{R3.4}{The current framework processes slices independently, which may lead to through-plane inconsistencies in volumetric reconstructions. This limitation is partially mitigated by the large inter-slice spacing of the datasets, where anatomical correspondence between adjacent slices is inherently weak.
Improving cross-slice coherence remains an important direction for future work. Given the explicit modeling of anatomical structure, natural extensions include enforcing smooth trajectories of latent anatomical codes along the slice direction or encouraging consistency in the canonical shape space across adjacent slices.}{}

\subsubsection{\rev{}{Computational Considerations}{}}

\rev{R3.4}{Our framework introduces additional computational and memory costs beyond standard encoder--decoder segmentation networks, primarily due to the anatomical bases, registration module, and multi-scale warping. This overhead is configurable through the number of bases and warping scales, allowing a flexible trade-off between accuracy and efficiency.
At the same time, our method avoids computationally intensive objectives such as adversarial training or explicit cross-domain feature alignment, and remains practical under typical GPU constraints. Further efficiency gains may be achieved through basis pruning or distillation of the learned anatomical representations.}{}

\subsubsection{\rev{}{Modeling Assumptions of the Semantic Manifold}{}}

\rev{R4.3}{
Our work models the bases using Gaussians to ensure stable
optimization and tractable inference, which may limit expressiveness for complex anatomical variability.
Exploring richer formulations, such as mixture-based priors or normalizing flows,
is a natural direction for future work, albeit with increased challenges in
identifiability, regularization, and optimization.
Moreover, we adopt a simple weighted
log-linear aggregation to compose bases conditioned on the blending vector $\w$, which may limit expressiveness in
highly heterogeneous anatomical settings.
More expressive aggregation mechanisms, such as cross-attention, can be integrated easily, at the cost of additional
computational and optimization complexity.
}{}

\subsubsection{\rev{}{Dataset Bias}{}}
\rev{R1.2,\\R2.2\\R4.2\\R4.9}{
As with most publicly available benchmark datasets, the evaluated datasets may
exhibit inherent biases related to cohort composition, acquisition protocols, distribution of anatomical variability, and annotation conventions, which
implicitly introduce data and label noise.
Our evaluation follows standard benchmark protocols without introducing additional
selection criteria or noise-specific assumptions, and thus reflects model behavior
under naturally occurring, non-ideal conditions, enabling fair comparison with
prior work.
Nevertheless, the disease spectrum and demographic diversity of these datasets may
not fully reflect real-world clinical populations, and validation on larger and more
diverse cohorts remains an important direction for future work.
}{}

\subsubsection{\rev{}{Extensions Beyond the Evaluated Setting}{}}
\rev{R1.1,\\R4.8}{
Although we focus on unsupervised adaptation, our method is not tied
to a specific supervision pattern, as it performs generative modeling
at the level of individual images:
each image contributes to learning through reconstruction,
while segmentation supervision, when available, is incorporated as an additional objective.
As a result, the framework naturally generalizes to more general settings,
including multi-source and partially annotated
datasets, where labeled and unlabeled images can be jointly incorporated into training.
A systematic evaluation of these extensions is left as future work.
Moreover, our framework is evaluated on
pixel-wise segmentation of structural medical images, where anatomical shape can be meaningfully modeled.
While the current instantiation focuses on MRI and CT segmentation, the underlying
principle of decomposing images into shared semantic knowledge and subject-specific
variations is not tied to a specific modality or task.
Extending the framework to other modalities (e.g., PET or ultrasound) or to
tasks like detection or classification would require
specific observation models and supervision schemes, and is left
as a promising direction for future work.}{}

\section{Conclusion}

We have proposed a unified and semantically grounded framework for unsupervised domain adaptation in medical image segmentation, which seamlessly supports both source-accessible and source-free scenarios. Our method explicitly disentangles canonical anatomy and individual-specific geometry through a shared latent manifold within a theoretically grounded Bayesian framework. By leveraging a structured composition of learnable anatomical bases, our method enables explainable \rev{R3.2}{and emergent}{} domain adaptation \rev{}{indirectly via a structured, shared semantic space}{that naturally emerges from the framework design}, without \rev{}{relying on explicit cross-domain alignment strategies}{the need for any handcrafted adaptation strategies}. Extensive experiments on public multi-organ and multi-modality benchmarks demonstrate state-of-the-art performance of our model, particularly in the highly challenging source-free setting, with strong generalization and robustness under various domain shifts. Beyond segmentation accuracy, we illustrate the strong interpretability of our framework by visualizing the disentanglement, manifold traversal, and domain alignment results. \rev{R4.3}{}{A limitation of this work is that the semantic manifold is modeled using relatively simple Gaussian distributions. Future work will explore richer formulations that could enhance adaptability in more complex clinical scenarios.}

%% file: _symbols.tex
\newcommand{\myeq}[1]{%
  \begin{equation}
\begin{alignedat}{2}
    #1
  \end{alignedat}
  \end{equation}
}

\newcommand{\red}[1]{\textcolor{red}{#1}}
\renewcommand{\mid}{\mathclose{}|\mathopen{}}
\newcommand{\ie}{\emph{i.e.}}
\newcommand{\eg}{\emph{e.g.}}
\newcommand{\defas}{\coloneqq}
\newcommand{\kl}[2]{D_{\text{KL}}\left[#1\ \|\ #2\right]}
\newcommand{\x}{\mathbf{x}}
\newcommand{\y}{\mathbf{y}}
\newcommand{\z}{\mathbf{z}}
\newcommand{\st}{\boldsymbol{\phi}}
\newcommand{\s}{\mathbf{s}}
\newcommand{\w}{\mathbf{w}}
\renewcommand{\v}{\mathbf{v}}
\renewcommand{\mid}{\mathclose{}|\mathopen{}}

\newcommand{\pw}{p(\w)}
\newcommand{\ps}{p(\s)}
\newcommand{\pv}{p(\v)}
\newcommand{\pvl}{p(\v^\lj)}
\newcommand{\pz}{p(\z|\w)}
\newcommand{\pzl}{p(\z^l|\w)}
\newcommand{\px}{p(\x|\z,\v,\s)}
\newcommand{\py}{p(\y|\z,\v)}
\newcommand{\qs}{q(\s|\x)}
\newcommand{\qw}{q(\w|\x)}
\newcommand{\qz}{q(\z|\w)}
\newcommand{\qzl}{q(\z^l|\w)}
\newcommand{\qv}{q(\v|\x,\z)}
\newcommand{\qvl}{q(\v^\lj|\x,\z^\lj,\v^{<\lj})}
\newcommand{\qvlessl}{q(\v^{<\lj}|\x,\z^{<\lj})}
\newcommand{\pall}{p(\x,\y,\w,\z,\v,\s)}
\newcommand{\qall}{q(\w,\z,\v,\s|\x,\y)}
\newcommand{\E}{\mathbb{E}}
\newcommand{\lj}{{l_j}}

\newcommand{\Lz}{\mathcal{L}_{\text{tem}}}
\newcommand{\Lznew}{\widetilde{\mathcal{L}}_{\text{tem}}}
\newcommand{\Lv}{\mathcal{L}_{\text{vel}}}
\newcommand{\N}[2]{\mathcal{N}(#1,#2)}
\newcommand{\mubold}{\boldsymbol{\mu}}
\newcommand{\Sigmabold}{\boldsymbol{\Sigma}}
\newcommand{\qmzl}{q_m(\z^l)}
\newcommand{\wdeterm}{\widetilde{\w}}
\newcommand{\wdetermm}{\widetilde{w}_m}
\newcommand{\sdeterm}{\mubold_s}
\newcommand{\Lalign}{\mathcal{L}_{\text{align}}}
\newcommand{\Lbalance}{\mathcal{L}_{\text{usage}}}
\newcommand{\Lwdice}{\mathcal{L}_{\text{struct}}}
\newcommand{\Lx}{\mathcal{L}_{\text{recon}}}
\newcommand{\Ly}{\mathcal{L}_{\text{seg}}}
\newcommand{\B}{\mathcal{B}}
\newcommand{\Bs}{\B^s}
\newcommand{\Bt}{\B^t}
\newcommand{\Qs}{\mathcal{Q}^s}
\renewcommand{\c}{\mathbf{c}}
\newcommand{\e}{\mathbf{e}}
\newcommand{\fr}[2]{D_{\text{FR}}\left[#1\ \|\ #2\right]}

\newcommand{\Lelbo}{\mathcal{L}_{\text{LB}}}
\newcommand{\mscmr}{MS-CMRSeg}
\newcommand{\amos}{AMOS22}

\newcommand{\circnew}{\,\raisebox{1pt}{\tikz \draw[line width=0.6pt] circle(1.2pt);}\,}